\documentclass[10pt,twocolumn,letterpaper]{article}

\usepackage{cvpr}              % To produce the CAMERA-READY version

\usepackage{tikz}
\usetikzlibrary{tikzmark}

\def\hlinewd#1{\noalign{\ifnum0=`}\fi\hrule \@height #1 \futurelet \reserved@a\@xhline}

\definecolor{brown1}{RGB}{227, 204, 194}  % 定义自定义蓝色（钢蓝色）
\definecolor{brown2}{RGB}{247, 219, 189}  % 定义自定义蓝色（钢蓝色）

\definecolor{myblue}{RGB}{42, 116, 174}  % 定义自定义蓝色（钢蓝色）
\definecolor{fullmarkcolor}{RGB}{214, 233, 213} % Light green for Full mark row
\definecolor{opensource}{RGB}{255, 243, 206}
\definecolor{opensource2}{RGB}{217, 231, 252}
\definecolor{streaming}{RGB}{214, 239, 244} % Light red for W/O videos row
\definecolor{closesource}{RGB}{184, 219, 179}
\definecolor{blue1}{rgb}{0.8, 0.9, 1.0}  % 浅蓝色
\definecolor{blue2}{rgb}{0.7, 0.85, 1.0}  % 中蓝色
\definecolor{blue3}{rgb}{0.4, 0.7, 0.9}  % 深蓝色

\definecolor{scifiBlue}{rgb}{0.2, 0.7, 1.0}   % 明亮的电子蓝
\definecolor{scifiRed}{rgb}{1.0, 0.2, 0.2}     % 炽热的科幻红
\definecolor{scifiGreen}{rgb}{0.4, 1.0, 0.3} % 备选：荧光绿
\definecolor{scifiPurple}{rgb}{0.6, 0.2, 0.8} % 备选：深邃的紫罗兰
\definecolor{demphcolor}{gray}{.5}

\usepackage{pifont}
\usepackage{enumitem}
\usepackage{adjustbox}
\usepackage[utf8]{inputenc} % allow utf-8 input
\usepackage[T1]{fontenc}    % use 8-bit T1 fonts
\usepackage{url}            % simple URL typesetting
\usepackage{booktabs}       % professional-quality tables
\usepackage{amsfonts}       % blackboard math symbols
\usepackage{nicefrac}       % compact symbols for 1/2, etc.
\usepackage{microtype}      % microtypography
\usepackage{xcolor}         % colors
\usepackage{colortbl} 
\usepackage{graphicx}
\usepackage{amsmath}
\usepackage{multirow} 
\usepackage{caption}
\usepackage{mathtools}
\definecolor{lightgold}{RGB}{255, 242, 217}
\usepackage{algorithm}
\usepackage{algpseudocode}

\usepackage{inconsolata} % Or any other monospaced font package
\usepackage{tcolorbox} % For creating nice boxes
\tcbuselibrary{skins}
\usepackage{array}

\definecolor{miragepink}{RGB}{255, 0, 127} % 定义一个名为 'miragepink' 的新颜色

\usepackage[table]{xcolor}
\usepackage{booktabs}
\definecolor{IncreaseColor}{HTML}{2E86AB} % blue for increase
\definecolor{DecreaseColor}{HTML}{06A77D} % green for decrease

\definecolor{cvprblue}{rgb}{0.21,0.49,0.74}
\usepackage[pagebackref,breaklinks,colorlinks,allcolors=cvprblue]{hyperref}

\hypersetup{
    colorlinks=true, % 启用颜色链接
    linkcolor=cvprblue, % 设置内部链接（如目录、图表引用）的颜色
    citecolor=cvprblue, % 设置参考文献引用的颜色
    urlcolor=cvprblue, % 设置 URL 链接的颜色
}

\def\paperID{} % *** Enter the Paper ID here
\def\confName{CVPR}
\def\confYear{2026}

\title{WeaveTime: Stream from Earlier Frames into Emergent Memory in VideoLLMs}
\author{
Yulin Zhang$^{1}$ \hspace{1.5em} Cheng Shi$^{3}$ \hspace{1.5em} Sibei Yang$^{2\dagger}$\\
{$^1$ShanghaiTech University}\hspace{1.5em}
{$^2$School of Computer Science and Engineering, Sun Yat-sen University} \\ {$^3$School of Computing and Data Science, The University of Hong Kong} \\
\textbf{Project Page: }\url{https://zhangyl4.github.io/publications/weavetime/}}

\begin{document}
\maketitle

\begin{abstract}
Recent advances in Multimodal Large Language Models have greatly improved visual understanding and reasoning, yet their quadratic attention and offline training protocols make them ill-suited for \emph{streaming} settings where frames arrive sequentially and future observations are inaccessible. 
We diagnose a core limitation of current Video-LLMs, namely Time-Agnosticism, in which videos are treated as an unordered bag of evidence rather than a causally ordered sequence, yielding two failures in streams: \emph{temporal order ambiguity}, in which the model cannot follow or reason over the correct chronological order, and \emph{past–current focus blindness} where it fails to distinguish present observations from accumulated history. 
We present \textbf{WeaveTime}, a simple, efficient, and model-agnostic framework that first \emph{teaches} order and then \emph{uses} order. We introduce a lightweight \textit{Temporal Reconstruction} objective—our \emph{Streaming Order Perception} enhancement—that instills order-aware representations with minimal finetuning and no specialized streaming data. At inference, a \textit{Past–Current Dynamic Focus Cache} performs uncertainty-triggered, coarse-to-fine retrieval, expanding history only when needed. 
Plugged into exsiting Video-LLM without architectural changes, WeaveTime delivers consistent gains on representative streaming benchmarks, improving accuracy while reducing latency. These results establish WeaveTime as a practical path toward time-aware stream Video-LLMs under strict online, time-causal constraints. Code and weights will be made publicly available.
\end{abstract}
    
\section{Introduction}
\label{sec:intro}

Modern visual intelligence systems are increasingly deployed in settings where the world is revealed not as a finished video, but as an online stream along the arrow of time. Frames arrive sequentially, events must be interpreted in the order they occur, the ``present'' is observed exactly once, the ``past'' accumulates into memory, and the ``future'' remains unavailable. Such strictly time-ordered streams underpin safety–critical and interactive applications, including autonomous driving~
\cite{wangOmniDriveHolisticLLMAgent2024, zhengDoe1ClosedLoopAutonomous2024}, human-robot interaction~\cite{kimOpenVLAOpenSourceVisionLanguageAction2024, florenceImplicitBehavioralCloning2021}, real-time surveillance, and online conferencing.

\begin{figure}[t] 
    \centering % Centers the content inside the figure
    \includegraphics[page=1,width=0.48\textwidth,trim=0 0 220 0,clip]{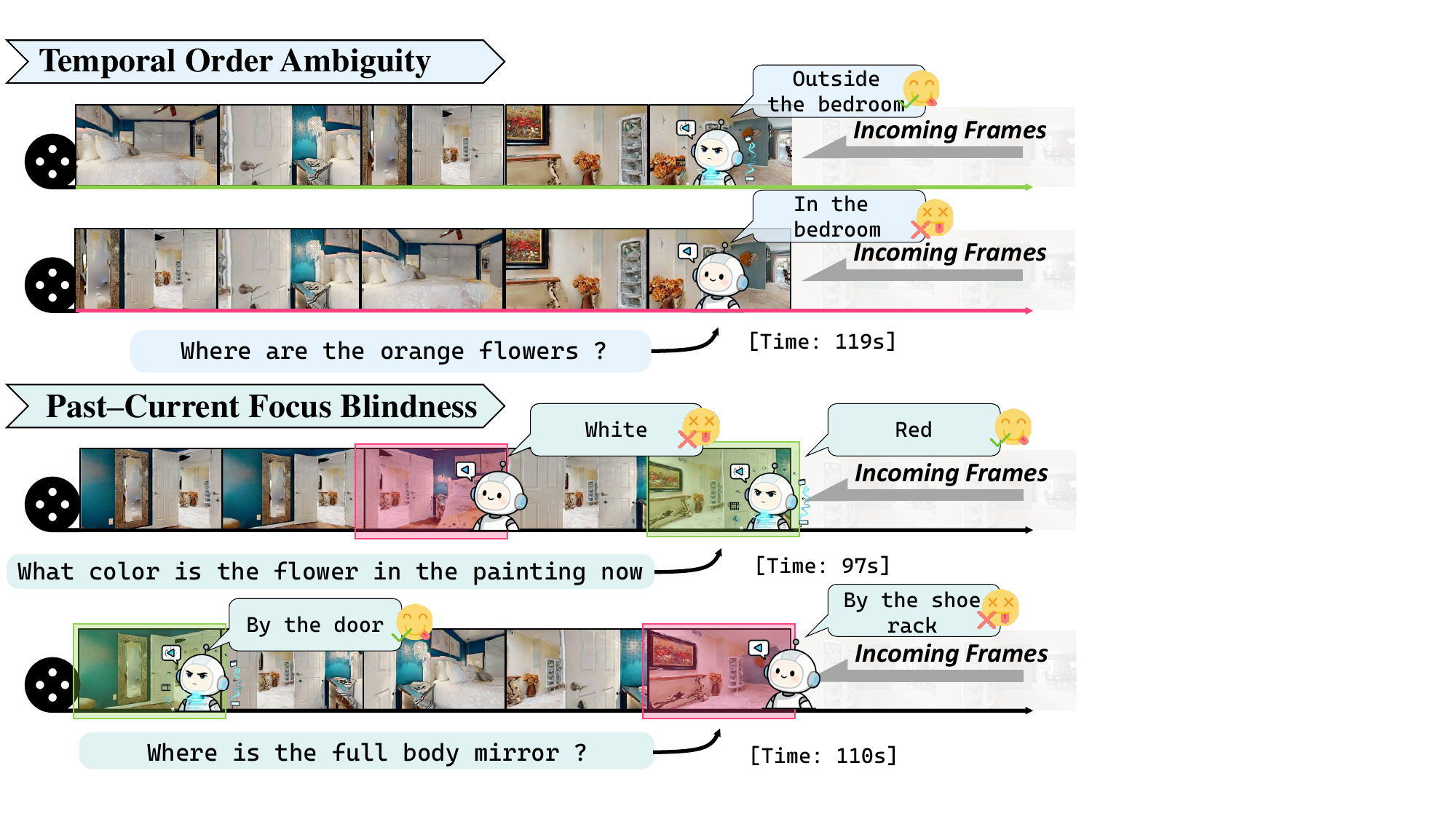} % <left> <bottom> <right> <top> Your beautiful image inclusion 
    \caption{\textbf{Illustrative examples of two coupled challenges in streaming Video-LLMs stemming from Time-Agnosticism. }Top: Temporal Order Ambiguity. The model struggles to correctly interpret the temporal sequence of events (e.g., entering vs. leaving a room), leading to erroneous spatial inferences (e.g., mislocating the orange flowers). Bottom: Past–Current Focus Blindness. The model fails to dynamically allocate attention between the immediate observation and relevant past memories. For the "What color is the flower in the painting now?" query, the answer is in the current frame, but the model recalls an irrelevant past moment. Conversely, for the "Where is the full body mirror?" query, which requires historical context, the model fixates on the current frame, leading to an incorrect answer.}
    \label{fig:example}
\end{figure}

To tackle these scenarios, recent work~\cite{chenVideoLLMonlineOnlineVideo2024, chenLiveCCLearningVideo2025, di2025rekv, wang2025streambridgeturningofflinevideo} begin to develop \emph{streaming} Video-LLMs that build on and benefit from the rapid progress of general Video-LLMs~\cite{wang2024qwen2vlenhancingvisionlanguagemodels, li2024llavaonevisioneasyvisualtask, li2024llavanextinterleavetacklingmultiimagevideo, yao2024minicpmvgpt4vlevelmllm}. These models reuse powerful Video-LLMs while augmenting them with stream-style training pipelines and memory mechanisms for online scene understanding~\cite{huang2025onlinevideounderstandingovbench,liOVOBenchHowFar2025, linStreamingBenchAssessingGap2024}. In this regime, \emph{time perception} becomes central: the model must represent not only \emph{what} happens but also \emph{when} it happens, consistently distinguishing accumulated past evidence from the currently visible scene.

% Continuous visual streams—such as those in autonomous driving, human–robot interaction, real-time surveillance, and online meetings—unfold under a strict arrow of time. Frames arrive sequentially; events evolve in order; the future is unavailable. Unlike offline video understanding, \emph{streaming} video understanding operates under online and time-causal constraints: at each step, the model must reason from the current observation while selectively drawing on an ever-growing memory of the past to answer questions that may arrive at any moment.

% Recent progress in Video-LLMs has been substantial, driven by powerful vision–language backbones and large-scale training data \cite{ref}. Streaming variants often build upon these foundations via (i) specialized data pipelines that synthesize stream-like inputs from static video datasets, and (ii) bespoke memory mechanisms designed to keep context within computational limits \cite{ref}. However, simply porting offline designs to the streaming regime overlooks a deeper mismatch: the model’s \emph{time perception}. In a stream, events possess an explicit developmental order and “past” must be meaningfully distinguished from “now”.

\begin{figure}[t]
    \centering
    \includegraphics[width=0.95\linewidth]{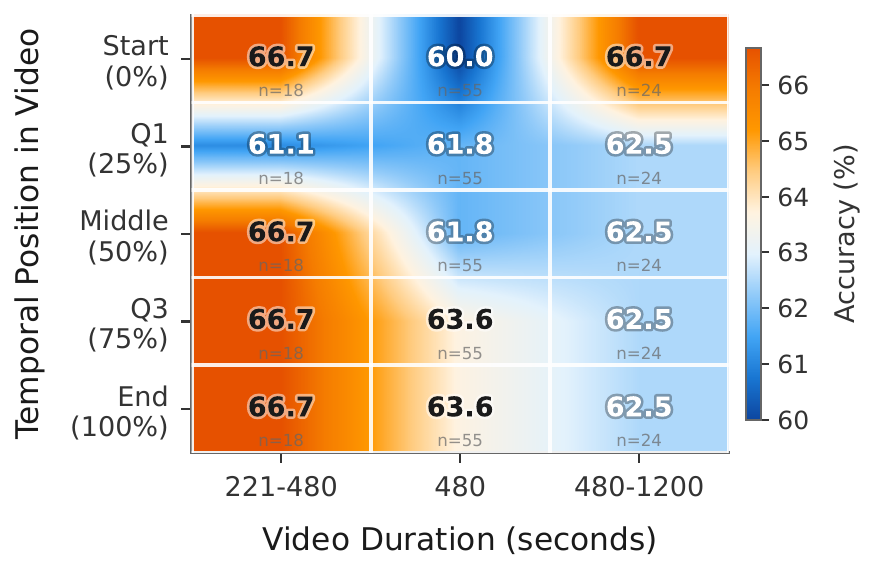}
    % \caption{\textbf{Impact of Ground Truth Answer Window Shift Position on Video QA Accuracy.} This heatmap illustrates the accuracy of the Video Question Answering (QA) model when the ground truth (GT) answer window is artificially shifted to different relative positions (0\% to 100\%) within the video. The accuracy is reported across videos grouped by their binned length (e.g., $345\text{s}, 480\text{s}$, etc.). The observed variation in accuracy with respect to the shift position and video length suggests the presence of a temporal positional bias in the model, indicating that the model may preferentially attend to or rely on information located at specific temporal locations within the video rather than consistently locating the relevant content based on the query.}
    \vspace{-0.2cm}
    \caption{\textbf{Impact of ground-truth answer-window shift on Video-QA accuracy.}
This heatmap shows model accuracy when the ground-truth (GT) answer window is shifted to different relative positions (0--100\%) along the video, evaluated over videos grouped by binned length (e.g., $345$\,s, $480$\,s).
Systematic changes in accuracy across shift positions and lengths reveal a temporal positional bias, whereby the model prefers specific time locations rather than consistently locating evidence based on the query.}

    \label{fig:position_bias}
    \vspace{-0.2cm}
\end{figure}

However, we empirically identify an inherent limitation of existing Video-LLMs~\cite{wang2024qwen2vlenhancingvisionlanguagemodels, li2024llavaonevisioneasyvisualtask} that we term \textbf{Time-Agnosticism}: the tendency to treat videos as an unordered \emph{bag of evidence} rather than a causally ordered temporal sequence. As shown in Tab.~\ref{tab:model_human_shuffle}, shuffling the frames barely harms the model and even improves accuracy on several tasks (highlighted in red), including explicitly temporal ones. In contrast, human performance collapses on temporal and action-related tasks once order is disrupted and only recovers when timestamps are provided. This discrepancy indicates that the model does not genuinely construct or exploit temporal order. Instead, as illustrated in Fig.~\ref{fig:position_bias}, it tends to focus on the beginning and end of short videos, while for longer videos it disproportionately attends to the initial segment. Consequently, the model relies on spatio-temporal shortcuts and positional biases~\cite{xing2024mitigatingobjecthallucinationconcentric,sharmaLosingVisualNeedles2024} that suffice for many benchmarks~\cite{fuVideoMMEFirstEverComprehensive2024,zhouMLVUComprehensiveBenchmark2024, wang2024lvbenchextremelongvideo} but fail to capture causal progression.

In streaming scenarios, the cost of Time-Agnosticism is magnified. We trace its impact to two coupled challenges:
\begin{itemize}[leftmargin=*, itemsep=0pt]
\item \textbf{Temporal Order Ambiguity.} A single query may relate to multiple semantically similar but differently ordered historical segments; correctness hinges on referencing them in the right order. When order is not encoded, attention drifts to temporally mismatched evidence. For instance, in the upper case of Fig.~\ref{fig:example}, the model’s confusion over temporal order leads it to misinterpret the action of ``leaving'' the room as ``entering'' it, thereby mistakenly inferring that the flowers outside the room are located inside. % 例如Fig~\ref{fig:example}中上面的case，由于模型在Temporal Order 上的疑惑，将走出房间错看成了走进房间，导致误认为房间外的花在房间中。
\item  \textbf{Past–Current Focus Blindness.} Some queries can be answered from the immediate observation; others require targeted retrospection. Biased or indiscriminate use of the growing memory treats all past content as equally relevant, causing the model to lose focus as the context expands. For example, in the lower cases of Fig.~\ref{fig:example}, the answer to the painting-related question is directly visible in the current frame, yet the model—affected by its Past–Current Focus Blindness—unnecessarily recalls an irrelevant past moment. Conversely, when the question concerns the position of the full body mirror, effective reasoning requires temporal retrospection, but the model remains fixated on the present observation. % 例如Fig~\ref{fig:example}中下面的两个case，关于画答案就在“眼前”，但模型受限于Past–Current的Focus Blindness，去回忆了过去错误时刻，而对于全身镜的位置需要进行记忆回溯，但是模型却关注着当前帧。
\end{itemize}

% Recent work 正是由于忽视了videoLLM在流场景的上述挑战，所以要么是通过专用的数据和繁重的训练“强行” 的构建一个streaming Video LLM存在巨大资源代价，还是通过定制记忆机制精细的找出最核心的视觉信息便于模型问答但效果不令人满意。

\begin{table}[t]
\centering
\resizebox{0.48\textwidth}{!}{
\begin{tabular}{lccccc}
    \toprule
    \textbf{Task} & Base 
                  & \multicolumn{2}{c}{Model Shuffle} 
                  & \multicolumn{2}{c}{Human Shuffle} \\
    \cmidrule(lr){3-4}\cmidrule(lr){5-6}
                  & 
                  & w/ TS  & w/o TS 
                  & w/ TS  & w/o TS \\
    \midrule
    Information Synopsis   & 0.80 & \cellcolor{red!10}0.80 & \cellcolor{red!10}0.80 &  1.00 & 0.40 \\
    Attribute Perception   & 0.40 & \cellcolor{red!10}0.40 & 0.20 &  1.00 &  0.60 \\
    Action Reasoning       & 0.60 & \cellcolor{red!10}0.80 & \cellcolor{red!10}0.60 &  1.00 & 0.00 \\
    % Object Reasoning       & 0.20 & \cellcolor{red!10}0.20 & \cellcolor{red!10}0.20 &  1.00 & 0.40 \\
    Spatial Reasoning      & 1.00 & \cellcolor{red!10}1.00 & 0.80 &  1.00 & 0.60 \\
    Object Recognition     & 0.80 & 0.60 & \cellcolor{red!10}0.80 &  1.00 & 0.40 \\
    Counting Problem       & 0.80 & \cellcolor{red!10}0.80 & 0.40 &  1.00 & 0.00 \\
    Temporal Perception    & 0.40 & \cellcolor{red!10}0.40 & \cellcolor{red!10}0.60 &  1.00 & 0.20 \\
    % OCR Problems           & 0.40 & \cellcolor{red!10}0.40 & \cellcolor{red!10}0.40 &  1.00 & 0.40 \\
    Action Recognition     & 0.60 & \cellcolor{red!10}0.60 & 0.40 &  1.00 & 0.20 \\
    Temporal Reasoning     & 0.60 & 0.40 & \cellcolor{red!10}0.60 &  1.00 & 0.20 \\
    \bottomrule
\end{tabular}
}
\caption{\textbf{Model vs.\ human performance under shuffled-frame inputs.}
Base is accuracy on ordered videos; other columns show accuracy with shuffled frames, with or without timestamps.
Humans remain reliable on non-temporal tasks but collapse on temporal/action tasks without timestamps and recover once timestamps are provided, indicating genuine use of temporal order.
The model, however, stays strong under shuffling and barely benefits from timestamps, revealing time-agnostic, bag-of-evidence behavior.
Cells where shuffling improves accuracy are highlighted in red.}
\label{tab:model_human_shuffle}
\end{table}

Recent studies~\cite{wang2025streambridgeturningofflinevideo, chenVideoLLMonlineOnlineVideo2024, di2025rekv, zhangFlashVStreamMemoryBasedRealTime2024} have largely overlooked the aforementioned challenges of applying VideoLLMs in streaming scenarios. As a result, existing approaches~\cite{chenVideoLLMonlineOnlineVideo2024, zhang2025eyeswideopenego, wang2025streambridgeturningofflinevideo, chenLiveCCLearningVideo2025} either rely on extensive specialized datasets and costly training procedures to forcibly construct a streaming-capable VideoLLM—incurring significant computational and data expenses—or adopt customized memory mechanisms~\cite{zhangFlashVStreamMemoryBasedRealTime2024, di2025rekv, kim2025infinipotvmemoryconstrainedkvcache} to selectively retrieve key visual information for question answering, yet their performance remains unsatisfactory.

To address these issues, we introduce \textbf{WeaveTime}, a simple, efficient, and Video-LLM-agnostic framework for streaming video question answering. 
WeaveTime follows a two-stage philosophy: first, \emph{teach} order; then, \emph{use} order. Concretely, we (i) make temporal dependencies explicit via a lightweight \emph{Streaming Order Perception} training enhancement, and (ii) perform uncertainty-aware, \emph{coarse-to-fine} temporal retrieval at inference to allocate attention between the present and the most relevant slices of the past.

First, to resolve \emph{Temporal Order Ambiguity}, we introduce a data-friendly auxiliary task, \emph{Temporal Reconstruction (TR)}. TR compels the model to encode inter-segment order and causal directionality by reconstructing or verifying the correct temporal arrangement of observed snippets. In practice, we simply append the ordering sub-question before the original QA query in a single conversation, allowing the LLM to reuse intermediate computations without introducing separate optimization stages. This upgrades memory from an unordered cache to an order-aware state without requiring dedicated streaming datasets or heavy retraining.
Second, once order perception is established, we deploy a \emph{Past–Current Dynamic Focus Cache (PCDF-Cache)}. Unlike methods that re-scan the entire history at every step, PCDF-Cache adheres to a ``\emph{look now, recall if needed}'' principle. We use prediction uncertainty (e.g., entropy) as a trigger: low-uncertainty queries proceed with forward inference using only the present; elevated uncertainty activates a coarse-to-fine retrieval module that progressively narrows the temporal search to the most pertinent segments. This yields time-efficient context expansion aligned to the query’s needs.

\noindent\textbf{Contributions.}
(1) We diagnose \emph{Time-Agnosticism} in current Video-LLMs and show that disrupting temporal order yields limited degradation, revealing reliance on spatio-temporal shortcuts rather than causal reasoning.
(2) We propose \textbf{WeaveTime}, a plug-and-play, Video-LLM-agnostic framework for streaming VQA that requires no specialized streaming data.
(3) We introduce \emph{Streaming Order Perception} via a lightweight \textbf{Temporal Reconstruction} auxiliary task, which instills order-aware representations with minimal finetuning.
(4) We design a \textbf{Past–Current Dynamic Focus Cache} that performs uncertainty-aware, coarse-to-fine temporal retrieval, enabling selective and efficient use of history under time-causal constraints.
(5) Extensive experiments conducted across multiple representative streaming benchmarks (including OVObench and Streaming-Bench) and applied to advanced Video-LLM backbone demonstrate that WeaveTime effectively reduces temporal order ambiguity, mitigates past–current focus blindness, and achieves significant improvements in streaming performance and efficiency over strong baselines.

\begin{figure*}[t] 
    \centering % Centers the content inside the figure
    \includegraphics[page=2,width=0.95\textwidth,trim=0 0 0 275,clip]{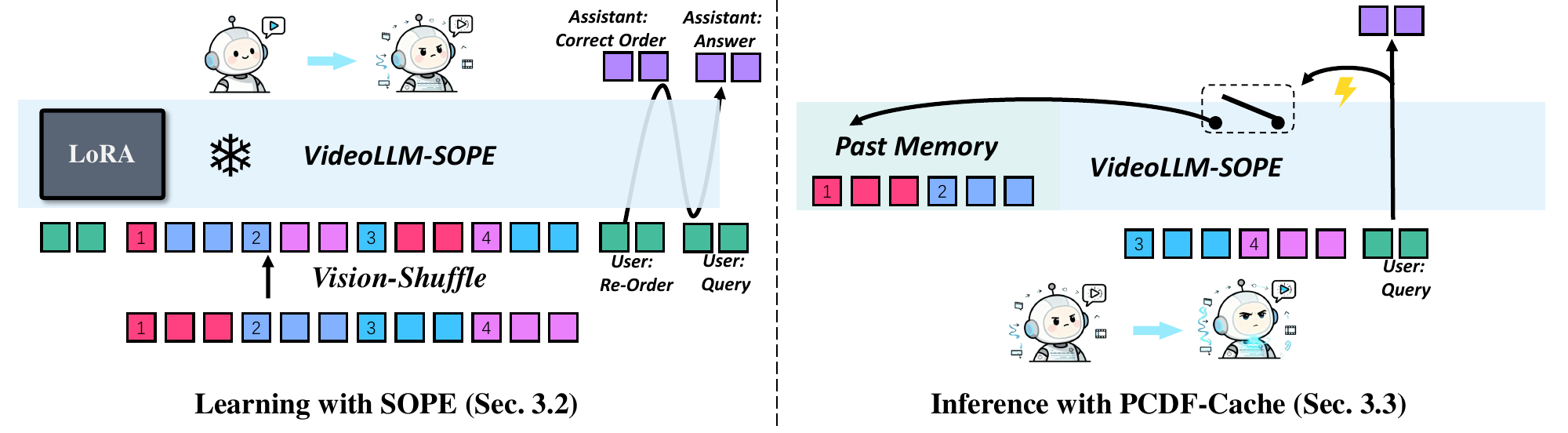} % <left> <bottom> <right> <top> Your beautiful image inclusion 你要两页还是1页 小小的也很可爱，这个训练是半栏得，inference是两栏得，两个图咋办 差两个图， trim改一下 why被你偷干净了 感觉没比裁剪轻松很多，是不是得在ppt里放好位置
    \caption{\textbf{Overview of WeaveTime.} (Left) The \emph{VideoLLM-SOPE} enhances temporal perception by reconstructing correct frame order from shuffled inputs during training. (Right) The \emph{Past–Current Dynamic Focus (PCDF) Cache} adaptively controls memory retrieval at inference, balancing between immediate observations and recalled past content. Together, these components jointly mitigate temporal ambiguity and inefficient memory access, enabling robust and temporally coherent streaming reasoning.}
    \label{fig:framework}
    \vspace{-0.3cm}
\end{figure*}

\section{Related Work}
\label{sec:rw}
\subsection{Video Large Language Models}
In recent years, Video Large Language Models have achieved remarkable progress in the field of video understanding. These models typically employ the ``Encoder-Projector-LLM'' architecture~\cite{liu2023visualinstructiontuning, li2024llavaonevisioneasyvisualtask, wang2024qwen2vlenhancingvisionlanguagemodels, li2024llavanextinterleavetacklingmultiimagevideo, deng3DLLaVAGeneralist3D2025}. 
The vast majority of existing Video-LLMs are designed for offline settings, assuming full accessibility to the entire video content and corresponding question. This design has demonstrated strong performance in short video tasks~\cite{liu2024tempcompassvideollmsreally, li2024mvbenchcomprehensivemultimodalvideo, pătrăucean2023perceptiontestdiagnosticbenchmark}.
In the context of long videos~\cite{yeMMEgoBuildingEgocentric2024, wang2024lvbenchextremelongvideo, ataallah2024infinibenchcomprehensivebenchmarklarge, zhouMLVUComprehensiveBenchmark2024}, models are constrained by the quadratic complexity of the attention mechanism~\cite{vaswani2023attentionneed}. Consequently, they commonly adopt a ``preprocessing paradigm'': the complete visual information is packaged into a fixed-length context via techniques like sampling~\cite{li2024llavaonevisioneasyvisualtask, li2024llavanextinterleavetacklingmultiimagevideo}, pruning~\cite{taoDyCokeDynamicCompression2024,huang2024prunevidvisualtokenpruning} or compression~\cite{shuVideoXLExtraLongVision2024, qianStreamingLongVideo2024, liuVideoXLProReconstructiveToken2025}.
While this "after-the-fact" strategy has yielded strong results on standard video benchmarks~\cite{yeMMEgoBuildingEgocentric2024, zhouMLVUComprehensiveBenchmark2024, mangalamEgoSchemaDiagnosticBenchmark2023, barmannWhereDidLeave2022, wang2024lvbenchextremelongvideo}, it is fundamentally unsuitable for streaming scenarios where complete visual content and queries are unavailable upfront. Furthermore, it inherently suffers from time-agnosticism, failing to properly perceive the sequential order of events and the correct distinction between the past and the future.
% 近年来，Video-LLMs~\cite{} 在视频理解领域取得了显著进展。这些模型通常采用“Encoder-Projector-LLM”架构 ~\cite{}。
% 现有的 Video-LLMs 绝大多数是为 offline settings 设计的，假设能“纵览”整个视频 (fully accessible)和对应的问题，which 已经表现优异在short video task中。
% 在长视频任务中，受限于注意力机制的二次复杂度 (quadratic complexity)，它们普遍采用一种**“预处理范式” (preprocessing paradigm)** [cite]： ，并通过采样 (sampling)~\cite{} 或压缩~\cite{} (compression)，将完整的视觉信息 "package" 到一个固定长度 (fixed-length) 的上下文中。
% 这种“事后诸葛亮”式的策略 (This "after-the-fact" strategy) 虽然achieve 了strong results on standard video benchmarks~\cite{}， 然而，在无法获取完整视觉内容和查询的流场景存在根本不适用，同时遭受着time-agnostic 对于事件发生的order，过去和未来无法正确感知。

\subsection{Streaming Video Understanding}
Unlike traditional offline video understanding tasks, which require processing the entire video before answering, Online Video Understanding aims to evaluate a  ability to respond to questions that arise at any time, utilizing past information and current observations from a sequential video stream input.
To benchmark the streaming video capabilities of existing Video-LLMs, Some works~\cite{wang2024videollmknowsspeakenhancing, zhang2025eyeswideopenego, qianDispiderEnablingVideo2025, chenLiveCCLearningVideo2025} focus on evaluating proactive interaction capabilities within a given scenario. In contrast, other benchmarks~\cite{liOVOBenchHowFar2025, huangOnlineVideoUnderstanding2024, linStreamingBenchAssessingGap2024, zeng2025streamforestefficientonlinevideo, yang2025svbenchbenchmarktemporalmultiturn} adopt a more generalized perspective for evaluation. These assess a diverse range of task categories, such as Real-time processing~\cite{liOVOBenchHowFar2025, linStreamingBenchAssessingGap2024}, Recollection ability~\cite{liOVOBenchHowFar2025}, Omni-modality integration~\cite{linStreamingBenchAssessingGap2024} and Diverse task categories involving static and dynamic targets~\cite{zeng2025streamforestefficientonlinevideo}.

To bridge the gap between Video-LLMs and general streaming tasks, considerable research effort has been dedicated
Many approaches~\cite{wang2025streambridgeturningofflinevideo, zhangFlashVStreamMemoryBasedRealTime2024,chenVideoLLMonlineOnlineVideo2024, wuVideoLLMMoDEfficientVideoLanguage2024, dingStreamMindUnlockingFull2025, chatterjeeMemoryefficientStreamingVideoLLMs2025} have focused on collecting specialized streaming Instruction-Tuning data using advanced VLM~\cite{yao2024minicpmvgpt4vlevelmllm, wang2024qwen2vlenhancingvisionlanguagemodels, geminiteam2024gemini15unlockingmultimodal, openai2024gpt4ocard} and LLM~\cite{deepseekai2025deepseekv3technicalreport, grattafiori2024llama3herdmodels} techniques to train models. While these methods achieve promising performance, they incur substantial costs.
Concurrently, existing stream memory management methods are generally categorized into two classes based on where memory is processed: compression-based (update-end optimization) and retrieval-based (load-end optimization).
Compression-based methods construct a fixed-length memory by meticulously selecting~\cite{yang2025streammemqueryagnostickvcache}, merging~\cite{wang2025streambridgeturningofflinevideo}, or dropping~\cite{kim2025infinipotvmemoryconstrainedkvcache} visual features when new frames exceed the Video-LLM's context limit. They measure importance and use structures (e.g., clustering~\cite{zhangFlashVStreamMemoryBasedRealTime2024}, forest~\cite{zeng2025streamforestefficientonlinevideo}) to fuse or discard redundant content. While offering higher response speed, these methods incur information loss and yield suboptimal performance.
Retrieval-based~\cite{di2025rekv} methods retain all visual memories externally. The most relevant memories are reloaded upon query by encoding the query/visual content using the LLM/external VLM~\cite{radford2021learningtransferablevisualmodels, zhai2023sigmoidlosslanguageimage} and similarity schemes. These methods offer a higher performance ceiling but suffer from slower response times due to the reload process and a Loss of Temporal Focus.

\section{Method}
\label{sec:method}

As illustrated in Fig.~\ref{fig:framework}, \textbf{WeaveTime} addresses streaming video question answering (Streaming VQA) through a synergistic design that couples \emph{training-time temporal order learning} with \emph{inference-time dynamic memory control}. We argue that \emph{temporal ambiguity} and \emph{inefficient memory access} are intrinsically intertwined: robust online video understanding emerges only when both the model’s temporal perception (learned during training) and its retrieval behavior (controlled at inference) are improved in concert.

\noindent\textit{Roadmap.}
We first review the Streaming VQA setting and a retrieval-based Video-LLM baseline (Sec.~\ref{sec:prelim}).
We then introduce \emph{Streaming Order Perception Enhancement (SOPE)}, which explicitly teaches the model to perceive and reason about temporal order (Sec.~\ref{sec:sope}).
Finally, we present \emph{Past-Current Dynamic Focus Cache (PCDF-Cache)}, an uncertainty-gated, coarse-to-fine memory controller that recalls long-term context only when needed (Sec.~\ref{sec:pcdf}).

\subsection{Preliminary}
\label{sec:prelim}

\subsubsection{Streaming Video-QA}
\paragraph{Task definition.}
In streaming VQA, video frames $f_1, f_2, \ldots, f_T$ arrive in temporal order.
When a question $q$ is issued at time $t$, the model must generate an answer $a_t$ \emph{causally}, i.e., using only the observed frames $\{f_1,\ldots,f_t\}$.
This requires not only understanding the current observation but also forming and querying a temporally consistent memory over past content.

\medskip
\noindent \textbf{Model formulation.}
Existing streaming memory approaches are broadly categorized as compression-based or retrieval-based.
We adopt a retrieval-based baseline due to its ability to avoid information loss while preserving temporal consistency.
Following a standard Video-LLM stack, we use a \emph{vision encoder}, a \emph{connector} (to map vision tokens into the language space), and an \emph{LLM} for multimodal generation.
A retrieval memory module is integrated into the LLM to load relevant historical context from a growing memory.

Concretely, during video stream encoding, we apply sliding-window attention to each incoming frame $f_t$ and produce key–value pairs $(K_t, V_t)$, which are appended to a memory buffer $\mathcal{M}_{t-1}$:
\begin{align}
(K_t, V_t) &= \mathrm{Encode}\!\left(\mathcal{M}_{t-1}[-C:],\, f_t\right), \nonumber\\
\mathcal{M}_t &= \mathrm{Append}\!\left(\mathcal{M}_{t-1}, (K_t, V_t)\right),
\label{eq:encode}
\end{align}
where $C$ denotes the local window length (tokens) used for causal conditioning.

Upon receiving a query $q$, a retrieval step selects the top-$K$ relevant frames from $\mathcal{M}_t$:
\begin{equation}
\mathrm{Load}(\mathcal{M}_t, q) = \mathrm{TopK}\!\left( \mathrm{Sim}(f_i^{v}, f^{q}) \right)_{i=1}^{|\mathcal{M}_t|},
\label{eq:load}
\end{equation}
where $f_i^{v}$ is the frame-level visual key (e.g., the average of a frame’s token keys), $f^{q}$ is the query key (e.g., the average over text tokens), and $\mathrm{Sim}(\cdot,\cdot)$ is cosine similarity.
The answer is then generated from the loaded context:
\begin{equation}
a_t = \mathrm{Answer}\!\big(\mathrm{Load}(\mathcal{M}_t, q),\, q\big).
\label{eq:answer}
\end{equation}

While retrieval-based methods avoid lossy compression, they face two issues in streaming:
(i) repeated long-range reloads degrade latency; and
(ii) \textit{Past–Current Focus Blindness}---difficulty allocating attention between recent observations and relevant past segments---leads to distractors.
%WeaveTime introduces (a) an uncertainty-gated \emph{look-now, recall-if-needed} policy to reduce redundant retrieval, and (b) a hierarchical (\emph{coarse-to-fine}) recall to pinpoint relevant context efficiently.

\subsection{Learning to Perceive Temporal Order in Streaming}
\label{sec:sope}

In this section, we address \emph{Temporal Order Ambiguity}—the tendency of models to treat visual inputs as a bag of evidence and attend to semantically similar yet temporally misordered segments. We introduce \textbf{Streaming Order Perception Enhancement (SOPE)}, a training-time scheme that injects explicit temporal priors and order supervision, endowing the model with robust temporal awareness that underpins effective inference-time retrieval.

\medskip
\noindent \textbf{Timestamped interleaving and an re-ordering objective.}
As shown in Fig.~\ref{fig:framework}, let the patchified video-token sequence be
\[
\mathbf{X} = \big[\tilde{\mathbf{v}}_{1,1}, \ldots, \tilde{\mathbf{v}}_{1,N_f},\,
\tilde{\mathbf{v}}_{2,1}, \ldots, \tilde{\mathbf{v}}_{2,N_f},\, \ldots \big],
\]
where $\tilde{\mathbf{v}}_{i,j}$ is the $j$-th vision token of frame $i$, and $N_f$ is the number of tokens per frame.
We prepend a timestamp token $\mathbf{ts}_i$ before each frame’s tokens and then shuffle the frame content while keeping the timestamps explicit:
\begin{equation}
\mathbf{X}'=\big[\mathbf{ts}_{1},\, \tilde{\mathbf{v}}_{2,1}, \ldots, \tilde{\mathbf{v}}_{2,N_f},\,
\mathbf{ts}_{2},\, \tilde{\mathbf{v}}_{1,1}, \ldots, \tilde{\mathbf{v}}_{1,N_f},\, \ldots \big].
\label{eq:timestamped}
\end{equation}
We augment the QA prompt with a short instruction: 
``\emph{These video segments are shuffled. List each segment's true time range.}'' 
The model is first required to recover the correct temporal order, and then to answer the original question.
Instead of realizing this through the addition of an external order prediction head, we leverage the Large Language Model's (LLM's) inherent capacity for rearranging and recalling input text. After providing linguistic cues that display the timestamps for adjacent visual content, we formulate the predicted order as a next-token prediction task to generate a structured response. This approach seamlessly integrates the temporal task with the subsequent question-answering pair. Consequently, this design avoids the need for extra modules or customized loss functions. Detailed prompt contents are provided in the Appendix.
% Instead of 通过add a 额外的order预测head 来realize this，在提供了timestamp 显示的语言提示对于相邻的vision content之后，我们顺应LLM天然对于输入文本天然的回看重组能力， 将预测的order作为一个next token prediciton任务，生成结构化的回答，自然的与后续的question answer pair集成在一起。因此，不需要增加额外的模块和定制的损失。详细prompt内容见附件。

This auxiliary task encourages a consistent, causal temporal manifold, effectively turning the past from an unordered cache into a structured chain. 
With such temporal awareness established at training time, inference-time retrieval can \emph{locate when} evidence occurred—not just \emph{what} occurred—leading to more precise temporal reasoning.

\begin{figure*}[t] 
    \centering % Centers the content inside the figure
    \includegraphics[page=3,width=0.95\textwidth,trim=0 0 50 220,clip]{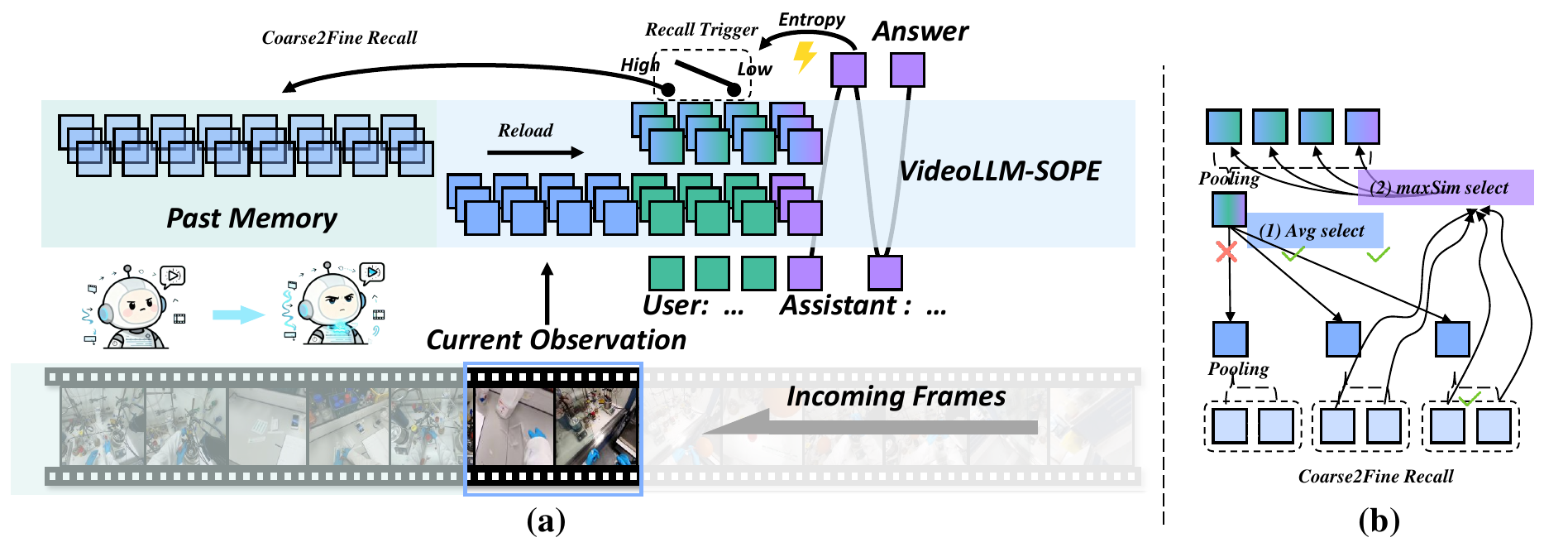} % <left> <bottom> <right> <top> Your beautiful image inclusion 你要两页还是1页 小小的也很可爱，这个训练是半栏得，inference是两栏得，两个图咋办 差两个图， trim改一下 why被你偷干净了 感觉没比裁剪轻松很多，是不是得在ppt里放好位置
    \caption{\textbf{Overview of the Past–Current Dynamic Focus Cache (PCDF-Cache).} (a) Given streaming inputs, the PCDF-Cache monitors prediction entropy: when uncertainty is low, the model answers directly from the current observation; when high, it triggers a recall from long-term memory via \emph{Coarse-to-Fine Recall}. (b) The recall module performs hierarchical selection—first pooling coarse candidates, then applying a max-similarity criterion for fine retrieval—balancing retrieval cost and contextual accuracy.}
    \label{fig:pcdfcache}
    \vspace{-0.3cm}
\end{figure*}

\subsection{Inference with Past-Current Dynamic Focus Cache}
\label{sec:pcdf}

With temporal order perception established at training time, the remaining challenge is inference-time control—determining whether to consult long-term memory and how to retrieve it efficiently. As shown in Fig.~\ref{fig:pcdfcache}, we propose \textbf{PCDF-Cache}, a lightweight controller that answers locally when confident, recalls globally when needed, and then applies coarse-to-fine refinement.

\subsubsection{Look Now, Recall if Needed}
PCDF-Cache implements an uncertainty-gated policy: use short-term context if confident; otherwise, recall.
When a query $q$ arrives at time $t$, the model first answers from the short window $\mathcal{M}_{t-1}[-C:]$:
\begin{equation}
a_t^{(0)} = \mathrm{Answer}\!\big(\mathcal{M}_{t-1}[-C:],\, q\big).
\label{eq:local-answer}
\end{equation}
We compute the predictive entropy $H_t=\mathrm{Entropy}\!\left(a_t^{(0)}\right)$ ~\cite{zou2025looktwiceanswermemoryspace} and compare it against a threshold $\delta$:
\begin{equation}
a_t =
\begin{cases}
a_t^{(0)}, & \text{if } H_t < \delta,\\[4pt]
\mathrm{Answer}\!\big(\mathrm{Load}_{\text{C2F}}(\mathcal{M}_t, q),\, q\big), & \text{otherwise.}
\end{cases}
\label{eq:gate}
\end{equation}
This uncertainty-gated recall reduces redundant long-range loads and mitigates Past–Current Focus Blindness by explicitly separating now from then.

\subsubsection{Coarse-to-Fine (C2F) Recall}
Exhaustive token-level retrieval is accurate yet intractable online. WeaveTime resolves this by first \emph{contracting} the search space with frame-level similarity (Eq.~\eqref{eq:load}) to form $\mathcal{M}_{\text{coarse}}$, then \emph{refining} via late-interaction, multi-vector matching~\cite{khattab2020colbert}. The result is token-level precision at a fraction of the latency and memory cost.

Let $\{f^{v}_{i,1},\ldots,f^{v}_{i,N_i}\}$ be the key vectors of the $i$-th frame’s vision tokens and $\{f^{q}_{1},\ldots,f^{q}_{N_q}\}$ be the key vectors of the query tokens.
We define a max-sim late-interaction score:
\begin{equation}
\mathrm{maxSim}\!\left(\{f^{v}_{i,k}\}, \{f^{q}_{j}\}\right) 
= \sum_{j=1}^{N_q} \max_{1 \le k \le N_i} \big\langle f^{q}_{j},\, f^{v}_{i,k} \big\rangle,
\label{eq:maxsim}
\end{equation}
and select the top-$K$ frames by this score within $\mathcal{M}_{\text{coarse}}$:
\begin{equation}
\mathrm{Load}_{\text{C2F}}(\mathcal{M}_t, q) 
= \mathrm{TopK}\!\left(
\underset{f_i \in \mathcal{M}_{\text{coarse}}}{\mathrm{maxSim}}
\!\left(\{f^{v}_{i,k}\}, \{f^{q}_{j}\}\right)
\right).
\label{eq:c2f}
\end{equation}

C2F recall avoids the heavy compute and memory footprint of one-shot fine-grained retrieval over the full history, while the late-interaction scoring provides robust, token-level cross-modal alignment for accurate selection.

\medskip
\noindent \textbf{Training--inference synergy.}
PCDF-Cache relies on the temporal awareness imparted by SOPE: timestamps and order cues help localize \emph{when} relevant events occurred, so that C2F recall searches in the right region and with the right granularity.
This coupling---order-aware training and uncertainty-gated, hierarchical recall at inference---forms the core of WeaveTime’s gains in both accuracy and responsiveness for streaming VQA.

\section{Experiments}
\label{sec:exp}

\subsection{Settings}
\label{sec:settings}

\noindent \textbf{Models and Baseline.}
We evaluate \textbf{WeaveTime} as a plug-in to representative open-source Video-LLM—LLaVA-OV-7B—to demonstrate generalizability without architecture modification. In addition, we consider both proprietary and open-source Video-LLMs, as well as a dedicated streaming model, as baselines. 
Following \emph{StreamBridge}, we adopt a multi-turn evaluation protocol tailored to streaming scenarios. Under this setting, we compare WeaveTime with two model-agnostic streaming enhancement methods: StreamBridge and ReKV. 

\medskip
\noindent \textbf{Implementation Details.}
For the Streaming Order Perception Enhancement (SOPE) training, we randomly sample $30$k offline video Instruction-Tuning (IT) data from the widely recognized, general synthetic video IT dataset LLaVA-Video-178K (which totals $1.3$M IT data); no additional streaming-specific data are utilized. We train for one epoch using LoRA with a learning rate of $1 \times 10^{-5}$. Training is conducted on $8$ GPUs.
For the Past–Current Dynamic Focus Cache (PCDF-Cache) inference, we leverage the ReKV codebase, implementing custom modifications for the cache. We also convert the data formats of Streaming-Bench and OVObench to conform to the ReKV interface. Following the experimental configuration of ReKV~\cite{di2025rekv}, we ensure a fair comparison by restricting the maximum number of recalled frames to $64$ during the retrieval process. Additionally, the Entropy threshold $\delta$ for our method is set to $0.6$.
% % For \emph{SOPE} training, we randomly sample 30k offline video IT 数据 from the 知名的，通用 synthetic video IT dataset LLaVA-Video-178K~\cite{ref} (共计 1.3M IT data), no additional streaming-specific data are used. We train 1 epoch with LoRA at a learning rate of 5e-5. Training is conducted on 8 \textsc{H20} GPUs.
% % For \emph{PCDF-Cache} inference， we use ReKV codebase，修改额外的cache实现，同时convert Streaming-Bench and OvO-Bench的data格式to 符合rekv的接口。
% 根据rekv的实验设置，我们在recall的时候也只召回64frame进行公平的对比，对于Entropy threshold $\delta$我们设置为0.6。

\definecolor{techblue}{RGB}{0, 102, 204}
\definecolor{techgray}{RGB}{240, 240, 240}
\definecolor{techdark}{RGB}{30, 30, 30}
\definecolor{techaccent}{RGB}{0, 153, 255}

\begin{table*}[htbp]
\centering
\huge
\setlength{\arrayrulewidth}{2pt}
\renewcommand{\arraystretch}{1.15}
\setlength{\tabcolsep}{6pt}
\resizebox{\textwidth}{!}{
\begin{tabular}{l|c|rrrrrrr|rrrrrrrrrrr}
\specialrule{3pt}{0pt}{0pt}  % 2pt 粗的横线
\textbf{Method} & \textbf{\# of Frames} &
\multicolumn{7}{c|}{\textbf{OVO-Bench Real-Time}} &
\multicolumn{11}{c|}{\textbf{Streaming-Bench Real-Time}} \\
\hline
 & & OCR & ACR & ATR & STU & FPD & OJR & AVG &
OP & CR & CS & ATP & EU & TR & PR & SU & ACP & CT & AVG \\
\hline
\rowcolor{techgray}
\multicolumn{20}{|c|}{\textbf{Human}} \\
\hline
Human & - &
93.96 & 92.57 & 94.83 & 92.70 & 91.09 & 93.20 & 91.30 &
89.47 & 92.00 & 93.60 & 91.47 & 95.65 & 92.52 & 88.80 & 88.75 & 89.74 & 91.30 & 91.46 \\
\hline
\rowcolor{techgray}
\multicolumn{20}{|c|}{\textbf{Proprietary Models (Offline), Single-Turn Evaluation}} \\
\hline
Gemini 1.5 pro \cite{geminiteam2024gemini15unlockingmultimodal} & 1 FPS &
85.91 & 66.97 & 79.31 & 58.43 & 63.37 & 61.96 & 69.32 &
79.02 & 80.47 & 83.54 & 79.67 & 80.00 & 84.74 & 77.78 & 64.23 & 71.95 & 48.70 & 75.69 \\
GPT-4o \cite{openai2024gpt4ocard} & 64 &
69.80 & 64.22 & 71.55 & 51.12 & 70.30 & 59.78 & 64.46 &
77.11 & 80.47 & 83.91 & 76.47 & 70.19 & 83.80 & 66.67 & 62.19 & 69.12 & 49.22 & 73.28 \\
\hline
\rowcolor{techgray}
\multicolumn{20}{|c|}{\textbf{Open-Source Models (Offline), Single-Turn Evaluation}} \\
\hline
Qwen2-VL-72B \cite{wang2024qwen2vlenhancingvisionlanguagemodels} & 64 &
65.77 & 60.50 & 69.83 & 51.69 & 69.31 & 54.35 & 61.92 &
- & - & - & - & - & - & - & - & - & - & - \\
LLaVA-Video-7B \cite{li2024llavanextinterleavetacklingmultiimagevideo} & 64 &
69.13 & 58.72 & 68.83 & 49.44 & 74.26 & 59.78 & 63.52 &
- & - & - & - & - & - & - & - & - & - & - \\
LLaVA-OV-7B \cite{li2024llavaonevisioneasyvisualtask} & 64/32 &
66.44 & 57.80 & 73.28 & 53.37 & 71.29 & 61.96 & 64.02 &
80.38 & 74.22 & 76.03 & 80.72 & 72.67 & 71.65 & 67.59 & 65.45 & 65.72 & 45.08 & 71.12 \\
Qwen2-VL-7B \cite{wang2024qwen2vlenhancingvisionlanguagemodels} & 64/1FPS &
60.40 & 50.56 & 66.03 & 47.19 & 66.34 & 55.43 & 55.98 &
75.20 & 82.81 & 73.19 & 77.45 & 68.32 & 71.03 & 72.22 & 61.39 & 61.47 & 46.11 & 69.04 \\
InternVL-V2-8B \cite{chen2024fargpt4vclosinggap} & 64/16 &
67.11 & 60.55 & 63.79 & 46.07 & 68.32 & 56.52 & 60.39 &
68.12 & 60.94 & 69.40 & 77.12 & 67.70 & 62.93 & 59.26 & 53.25 & 54.96 & 56.48 & 63.72 \\
\hline
\rowcolor{techgray}
\multicolumn{20}{|c|}{\textbf{Open-Source Models (Streaming), Single-Turn Evaluation}} \\
\hline
Flash-VStream-7B \cite{zhangFlashVStreamMemoryBasedRealTime2024} & 1 FPS &
24.16 & 29.36 & 28.45 & 33.71 & 25.74 & 28.80 & 28.37 &
43.59 & 25.87 & 24.91 & 23.87 & 27.33 & 13.08 & 18.52 & 25.20 & 23.87 & 48.70 & 23.23 \\
VideoLLM-Online-8B \cite{chenVideoLLMonlineOnlineVideo2024} & 2 FPS & 8.05 &
23.85 & 12.07 & 14.04 & 45.54 & 21.20 & 20.79 & 39.07 &
40.06 & 34.49 & 31.05 & 45.96 & 32.48 & 31.40 & 43.16 & 42.49 & 27.89 & 35.99  \\
VideoLLM-EyeWO-8B \cite{zhang2025eyeswideopenego} & 2 FPS &24.16 &27.52 &31.89 &32.58 &44.55 &35.87 &32.76 & - & - & - & - & - & - & - & - & - & - & -\\
Dispider \cite{qianDispiderEnablingVideo2025} & 1 FPS &
57.72 & 49.54 & 62.07 & 44.94 & 61.39 & 51.63 & 54.55 &
74.92 & 75.53 & 74.10 & 73.08 & 74.44 & 59.92 & 76.14 & 62.91 & 62.16 & 45.80 & 67.63 \\
\hline
\rowcolor{techgray}
\multicolumn{20}{|c|}{\textbf{Open-Source Models (Offline $\rightarrow$ Streaming), Multi-Turn Evaluation}} \\
\hline
\multicolumn{20}{l}{\cellcolor{scifiBlue!30}\textit{LLaVA-OV-7B}} \\
 + StreamBridge~\cite{wang2025streambridgeturningofflinevideo} & 1 FPS &
58.39 & 59.63 & 69.82 & 44.38 & 76.23 & 61.41 & 61.64 &
76.84 & 77.17 & 82.60 & 75.25 & 64.15 & 64.17 & 75.00 & 61.38 & 61.19 & 46.11 & 68.39 \\
 + ReKV$^\dagger$~\cite{di2025rekv} & 1 FPS &63.09 &55.05 &72.41 &46.63 &72.28 &60.87 &61.72 &65.85 & 78.12 & 77.92 & 71.84 & 66.88 & 65.11 & 69.44 & 62.20 & 61.08 & 43.09 & 66.15 \\ 
 + WeaveTime (Ours) & 1 FPS &72.48 & 69.72 & 74.13 & 53.37 &75.24 &67.93 & 68.82 & 71.54 & 81.25 & 86.75 & 78.64 & 75.16 & 73.21 & 72.22 & 69.11 & 68.75 & 44.68 & 72.13 \\
\multicolumn{20}{l}{\cellcolor{scifiGreen!30}\textit{Qwen2-VL-7B}} \\
 + StreamBridge~\cite{wang2025streambridgeturningofflinevideo} & 1 FPS &65.10 &64.22 &64.66 &46.63 &74.26 &65.22 &63.35 &80.38 &78.74 &83.22 &79.86 &74.21 &69.47 &77.78 &63.41  & 69.97 &43.01 &72.01 \\
 + ReKV$^\dagger$~\cite{di2025rekv} & 1 FPS  & 60.40 & 52.29 & 68.10 & 43.26 & 72.28 & 61.96 & 59.72 &71.00 & 82.03 & 80.13 & 74.76 & 71.34 & 69.16 & 73.15 & 64.63 & 68.75 & 45.74 & 70.07

 \\
 + WeaveTime (Ours) & 1 FPS 
&75.17 & 59.63 & 71.55 & 51.69 & 72.28 & 67.39 & 66.28  & 74.80 & 85.94 & 88.01 & 81.23 & 77.71 & 76.32 & 82.41 & 63.82 & 74.15 & 49.47 & 75.39 \\

\specialrule{3pt}{0pt}{0pt}  % 2pt 粗的横线
\end{tabular}
}
\caption{Comparison of various Video LLMs on \textbf{OVO-Bench Real-Time} and \textbf{Streaming-Bench Real-Time}. $^\dagger$ indicates experimental results derived from integrating the respective dataset into the ReKV\cite{di2025rekv} codebase.}
\label{tab:ovo_streaming_comparison_fixed}
\vspace{-0.2cm}
\end{table*}

\medskip
\noindent \textbf{Benchmarks.}
For multi-turn real-time understanding, we follow existing work~\cite{zhang2025eyeswideopenego,qianDispiderEnablingVideo2025} and evaluate on two representative streaming datasets: (1) OVO-Bench~\cite{liOVOBenchHowFar2025}, which comprises three subsets focusing on past, current, and future temporal cues, totaling twelve question types widely used to assess a Video-LLM's perception and understanding of various timelines in streaming scenarios; and (2) the Real-time subset of StreamingBench~\cite{linStreamingBenchAssessingGap2024}, which is dedicated to evaluating a model's real-time perception and comprehension in a streaming context, covering ten question categories.

To rigorously assess generalization, our ablations additionally include established offline benchmarks. Following the methodology in~\cite{di2025rekv}, we use the multi-choice subset of QAEgo4D~\cite{graumanEgo4DWorld30002022} and MLVU~\cite{zhouMLVUComprehensiveBenchmark2024} to focus on question-answering in long-form video understanding for Video-LLMs. Furthermore, to evaluate the streaming model's recall and event perception capabilities, we utilize the binary-choice question subset of EventHALL~\cite{baiSequentialModelingEnables2023}, which is designed to assess event hallucination in Video-LLMs.
% For multi-turn real-time understanding,  we follwing~\cite{} evaluate on two representative datasets, （1）OVO-Bench~\cite{ref}由包含对于过去，当前，未来的三个subset组成共计十二种问题类型，被广泛用于评估videoLLM再流场景中对于不同时间线索的感知和理解能力，and StreamingBench~\cite{ref} real-time 是 streamingbench中用于评估模型再流场景中实时感知和理解的subset，包含10类问题. 
% To assess generalization, our ablations additionally include offline benchmark such as QAEgo4D, MLVU, follow~\cite{di2025rekv} 我们使用，multi-choice subset focusing on question-answering in long long-form video understanding for videoLLM.
% and EventHALL 是用于评估videoLLM在视频中的事件幻觉，我们使用它的we use the binary-choice question，来评估流模型进行回忆和事件感知的能力.

\subsection{Main Results}
% \noindent \textbf{Streaming Real-Time Understanding.} 
Table~\ref{tab:ovo_streaming_comparison_fixed} presents the performance comparison of various Video-LLMs on \textbf{OVO-Bench Real-Time} and \textbf{Streaming-Bench Real-Time}. WeaveTime consistently outperforms all baseline models in multi-turn evaluation setting, showing a significant improvement in key metrics. Specifically, when integrated with LLaVA-OV-7B, WeaveTime achieves up to a \textbf{+7.10\%} improvement on OVO-Bench Real-Time and \textbf{+3.74\%} on Streaming-Bench Real-Time over StreamBridge and ReKV. 

In particular, notable improvements are observed in temporal-aware tasks such as Action Perception (ACP), Event Understanding (EU), and Action Recognition (ACR). Specifically, WeaveTime surpasses the baselines by +7.56\% in ACP, +9.04\% in EU, and +11.09\% in ACR. These results strongly demonstrate that our model's enhanced temporal perception and dynamic memory addressing capability significantly boost real-time video understanding.

% \noindent \textbf{General Video Understanding.} \todo{ChronoSense also performs robustly on conventional video understanding benchmarks, ensuring that our streaming optimizations do not compromise general capabilities. On the standard video QA tasks, the ChronoSense-augmented models show a \textbf{+3.5\%} increase in overall accuracy compared to the baseline LLaVA-OV-7B and Qwen2-VL-7B models. For instance, on the VQA2 dataset, ChronoSense achieves a top-1 accuracy of \textbf{82.3\%} versus \textbf{78.8\%} for the baseline, indicating that the improvements made for streaming tasks also translate to better general video comprehension.}

\subsection{In-Depth Analysis}
\label{sec:analysis}

\noindent \textbf{Ablation Study.}
Table~\ref{tab:tpe_tradeoff} reports a stepwise ablation that isolates the effect of each WeaveTime component. We begin with a strong Video-LLM baseline equipped with a naive retrieval-based memory~\cite{di2025rekv}. 

(1) Directly finetuning this baseline on a small amount of offline video (with timestamp prompts) degrades streaming performance due to distribution mismatch (-3.68\%). 

(2) Adding our \emph{Streaming Order Perception Enhancement} (SOPE) via the Temporal Reconstruction auxiliary task substantially improves accuracy over the naive finetune and surpasses the original baseline, indicating that SOPE effectively resolves Temporal Order Ambiguity in a data-friendly manner (+5.82\%) under the same data budget). 

(3) Finally, attaching the \emph{Past--Current Dynamic Focus Cache} (PCDF-Cache) further boosts performance by mitigating Past--Current Focus Blindness: the model adheres to a ``look now, recall if needed'' protocol and only expands temporal context when uncertainty is high, yielding additional gains while keeping memory/latency under control. 

% Overall, the progression
% \emph{Baseline} $\rightarrow$ \emph{Baseline+SOPE} $\rightarrow$ \emph{Baseline+SOPE+PCDF}
% consistently improves streaming VQA under time-causal constraints.

\definecolor{techgray}{RGB}{240, 240, 240}

\begin{table}[t]
\small
\centering
\setlength{\arrayrulewidth}{1pt}
\renewcommand{\arraystretch}{1.0}
\scriptsize
\resizebox{0.5\textwidth}{!}{
\begin{tabular}{c c c | c c | c c}
\specialrule{1pt}{0pt}{0pt}
\multicolumn{2}{c}{{SOPE Training}} &
{PCDF} &
\multicolumn{2}{c|}{{OVO-Bench}} &
\multicolumn{2}{c}{{Streaming-Bench}} \\
w/ TP & w/ TR & Cache & Overall & $\Delta$ & Real-Time & $\Delta$ \\
\specialrule{0.5pt}{0pt}{0pt}
\multicolumn{7}{l}{\cellcolor{scifiBlue!30}\textit{LLaVA-OV-7B + ReKV}} \\

 &  &  & 53.56 & -- & 66.15 & -- \\

\textcolor{scifiBlue}{\ding{52}} &  &  & 49.88 & \textcolor{red}{-3.68} & 65.91 & \textcolor{red}{-0.54} \\

\textcolor{scifiBlue}{\ding{52}} & \textcolor{scifiBlue}{\ding{52}} &  & 55.70 & \textcolor{green!50!black}{+5.82} & 68.49 & \textcolor{green!50!black}{+2.58} \\

\textcolor{scifiBlue}{\ding{52}} & \textcolor{scifiBlue}{\ding{52}} & \textcolor{scifiBlue}{\ding{52}} & {57.57} & \textcolor{green!50!black}{+1.87} & {72.13} & \textcolor{green!50!black}{+3.64} \\

\specialrule{0.5pt}{0pt}{0pt}
\end{tabular}
}
\vspace{-0.2cm}
\caption{{Ablation on SOPE Training and PCDF-Cache.} 
Performance is reported across OVO-Bench and the real-time subset of Streaming-Bench. {TP} refers to training that incorporates a small amount of offline video data augmented with timestamp tokens. {TR} denotes the inclusion of a temporal re-ordering auxiliary task. 
%$\Delta$ columns indicate the performance change from the previous configuration (\textcolor{green!50!black}{improvement} / \textcolor{red}{decrease}).
}
\label{tab:tpe_tradeoff}
\vspace{-0.3cm}
\end{table}

\medskip
\noindent \textbf{Effect of Coarse-to-Fine Recall.}
To assess our Coarse-to-Fine (C2F) temporal recall, we evaluate on long, egocentric videos with annotated temporal windows (\textsc{QAEgo4D}), a broad general-video benchmark (\textsc{MLVU}), and a hallucination-sensitive event dataset (\textsc{EventHALL}); results are summarized in Table~\ref{tab:retrieval_datasets}.
Compared with \emph{coarse-only} retrieval, C2F improves both recall and downstream task metrics by progressively narrowing the search to the most relevant time slices.
Compared with \emph{fine-only} retrieval, C2F avoids the ``VRAM wall'' by first filtering with a lightweight coarse pass and invoking fine-grained matching only when necessary.
In practice, C2F reduces the average number of retrieved frames/tokens and stabilizes latency, while maintaining higher answer accuracy across spatial and temporal queries.
These findings confirm that uncertainty-triggered, hierarchical retrieval is an efficient and effective way to allocate temporal attention in streaming settings.

\definecolor{techgray}{RGB}{245, 245, 245}
\definecolor{techblue}{RGB}{0, 102, 204}

\begin{table}[htbp]
\centering
\footnotesize
\renewcommand{\arraystretch}{1.2}
\resizebox{0.5\textwidth}{!}{
\begin{tabular}{lcccc}
\specialrule{1pt}{0pt}{0pt}  % 2pt 粗的横线
\multirow{2}{*}{{Method}} 
& \multicolumn{2}{c}{{QAEGO4D}} 
& \multicolumn{1}{c}{{MLVU}} 
& \multicolumn{1}{c}{{EVENTHALL}} \\

& {Recall $\uparrow$} & {Acc $\uparrow$} 
& {Acc $\uparrow$} 
& {Acc $\uparrow$} \\
\hline
{LLaVA-OV} & 14.0 & 52.8 & 64.7 & 60.1  \\
{+ ReKV} & 23.9 & 54.3 & 68.5 & 60.6  \\
\rowcolor{scifiBlue!30}
{+ C2F (Ours)} & {25.2}  & {55.2} & {68.9} & {61.4}  \\
{+ Fine} & \multicolumn{4}{c}{{OOM}} \\
\specialrule{1pt}{0pt}{0pt}  % 2pt 粗的横线
\end{tabular}
}
\vspace{-0.3cm}
\caption{{Retrieval strategy comparison across datasets.}
Each dataset reports Recall and/or Accuracy; “OOM” indicates out-of-memory.}
\label{tab:retrieval_datasets}
\vspace{-0.3cm}
\end{table}

\begin{figure}[t]
    \centering
    \includegraphics[width=1.0\linewidth]{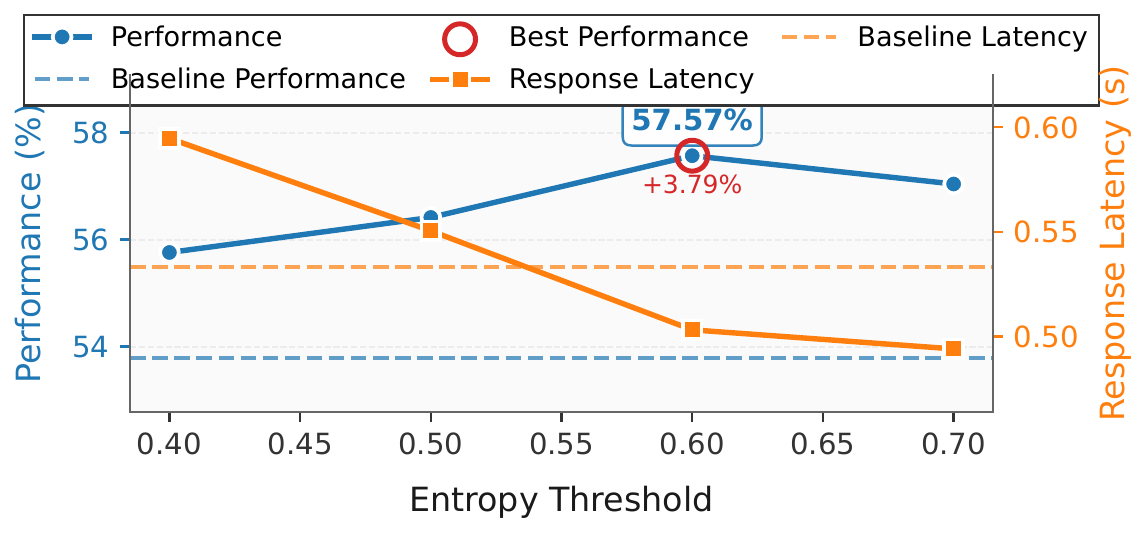}
    % \caption{\textbf{Ablation on the entropy threshold in PCDF-Cache.} Evaluated on OvO-Bench~\cite{liOVOBenchHowFar2025}, the threshold governs when memory recall is triggered during streaming QA. Accuracy (blue) peaks at 0.6, reflecting the need to balance current observation and past memory, while response latency (green) decreases with higher thresholds, confirming that adaptive recall yields the best trade-off between accuracy and efficiency.}
    \vspace{-0.4cm}
    \caption{\textbf{Ablation on the entropy threshold in PCDF-Cache.}
On OvO-Bench~\cite{liOVOBenchHowFar2025}, the threshold controls when memory recall is triggered.
Accuracy (blue) peaks at 0.6, balancing current observation and past memory, while response latency (green) decreases with larger thresholds, yielding the best accuracy–efficiency trade-off.}

    \label{fig:ab_gate}
    \vspace{-0.3cm}
\end{figure}

\medskip
\noindent \textbf{Analysis of PCDF-Cache.}
% As shown in Fig.~\ref{fig:ab_gate}, we ablate the Entropy Threshold in PCDF-Cache on the full OvO-Bench~\cite{liOVOBenchHowFar2025}. This threshold governs when the model triggers memory recall during streaming QA. We report overall accuracy (blue) and response latency (green), where latency denotes the time from question input to final answer.
% Accuracy first rises and then slightly declines as the threshold increases, peaking at 57.57\% at 0.6. This reflects the trade-off between leveraging current observations and retrieving past context: overly frequent recall introduces noise from outdated frames, while overly conservative recall weakens temporal grounding. Latency, in contrast, decreases monotonically, showing that unnecessary long-range reloads impose substantial overhead. By adaptively regulating recall, PCDF-Cache achieves a balanced trade-off between temporal reasoning performance and streaming efficiency.
As shown in Fig.~\ref{fig:ab_gate}, we conduct an ablation study on the full OvO-Bench~\cite{liOVOBenchHowFar2025} to analyze the effect of the Entropy Threshold hyperparameter in PCDF-Cache. The threshold controls when the model decides to trigger memory recall during streaming question answering. We measure both the overall accuracy (blue line) and the response latency (green line), defined as the elapsed time between question input and final response completion. 
In the blue curve, performance first increases and then slightly decreases as the entropy threshold rises, reaching its peak at 57.57\% when the threshold is set to 0.6. This trend highlights the importance of balancing attention between current observations and past memories—too frequent recall may cause distraction by outdated information, while overly conservative recall leads to insufficient temporal grounding. In the green curve, the response latency decreases monotonically as the entropy threshold grows. This indicates that recalling memory for every query is inefficient, as repeated long-range reloads introduce significant computational overhead and delay the response generation. By adaptively controlling the recall decision, PCDF-Cache achieves an optimal trade-off between temporal reasoning performance and streaming efficiency.

\begin{figure}[t] 
    \centering % Centers the content inside the figure
    \includegraphics[page=4,width=0.48\textwidth,trim=0 0 350 175,clip]{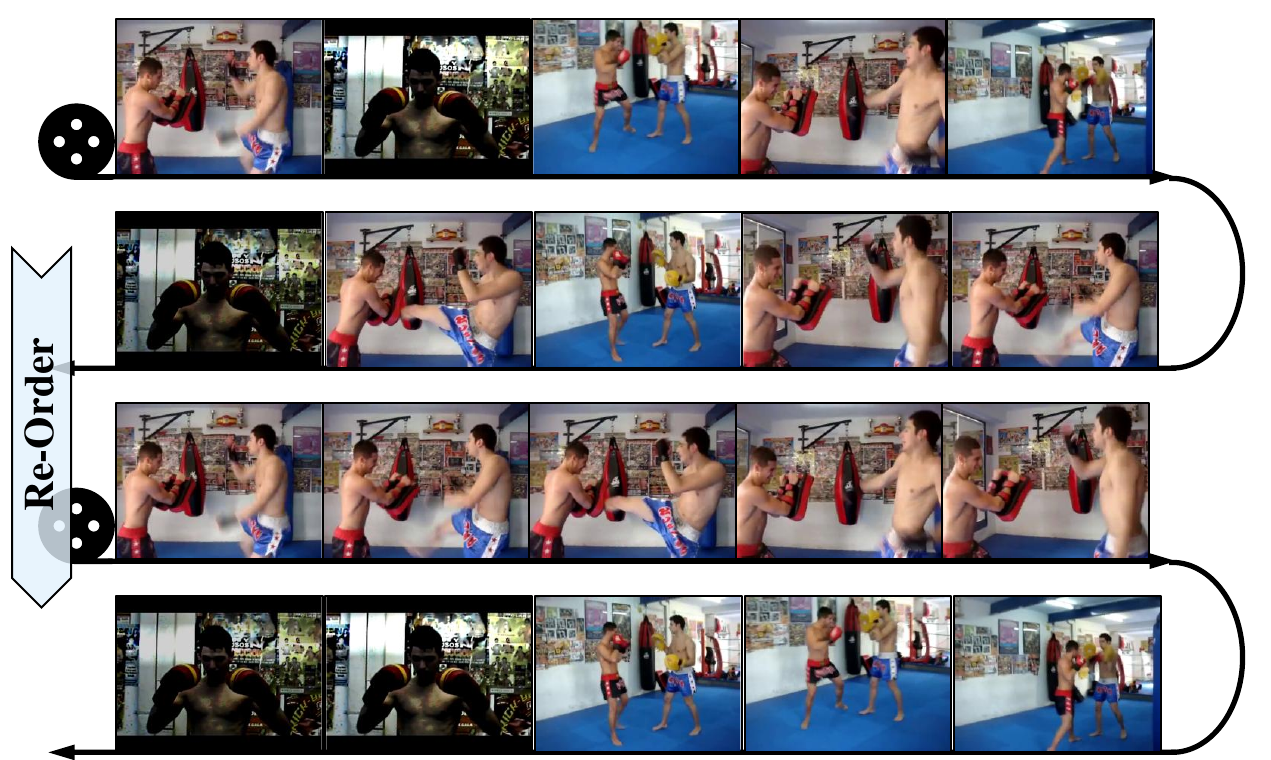} % <left> <bottom> <right> <top> Your beautiful image inclusion 
    \caption{
        \textbf{Example of Temporal Re-Ordering .} Visualization of the model's ability to restore chronological order. The top rows show a temporally-scrambled input sequence, which the model re-orders to produce the coherent output sequence in the bottom rows.
    }
    \label{fig:reorder_example}
    \vspace{-0.3cm}
\end{figure}

\begin{figure}[t]
    \centering
    \includegraphics[width=1.0\linewidth]{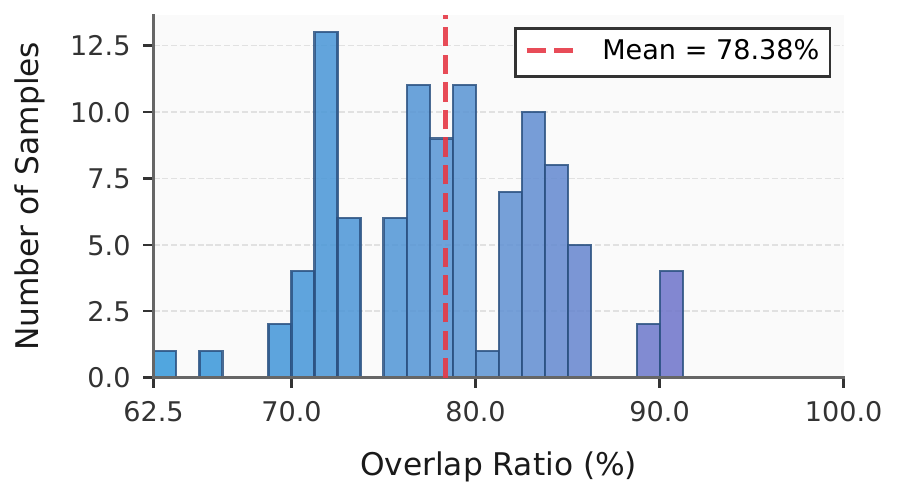}
    \caption{
        \textbf{Distribution of Temporal Re-order Overlap.} Histogram illustrating the model's performance distribution on the temporal re-ordering task.
    }
    \label{fig:reorder_rate}
    \vspace{-0.2cm}
\end{figure}

\medskip
\noindent \textbf{Analysis of the Feasibility of the Temporal Re-Order Task.}
To assess whether the model truly acquires the ability to perform temporal re-ordering, we sample 100 cases and evaluate the model’s predictions using next-token prediction in parallel. We then compare the predicted order with the ground truth to compute accuracy. As shown in Fig.~\ref{fig:reorder_rate}, after training, the model can reliably reconstruct the correct order of shuffled videos when guided by contextual cues. For clips that remain broadly coherent and preserve several key temporal anchors, the model achieves notably high reconstruction accuracy. For example, in Fig.~\ref{fig:reorder_example}, the model successfully re-orders a continuous action sequence, demonstrating that it has learned meaningful temporal structure rather than relying solely on static appearance cues.

\definecolor{techgray}{RGB}{240, 240, 240}

\begin{table}[t]
\small
\centering
\setlength{\arrayrulewidth}{1pt}
\renewcommand{\arraystretch}{1.25}
\scriptsize
\resizebox{0.45\textwidth}{!}{
\begin{tabular}{c | c c c}
\specialrule{1pt}{0pt}{0pt}
\multicolumn{1}{c|}{\textbf{Method}} & {OVO-Bench} & {GPUs} & {stream-tailored data} \\
\specialrule{0.5pt}{0pt}{0pt}
\multicolumn{1}{l}{\cellcolor{scifiBlue!30}\textit{LLaVA-OV + ReKV}} & 53.6 & -- & -- \\
\text{(+SOPE)} & 55.7 & 8 & 0 \\
\specialrule{0.5pt}{0pt}{0pt}
\multicolumn{1}{l}{\cellcolor{techgray}\textit{StreamForest Base}}  & 53.9 & 32 & -- \\
\text{(+Online IT)} & 55.6 & 32 & 121K \\
\specialrule{1pt}{0pt}{0pt}
\end{tabular}
}
\caption{
\textbf{Data and Compute Efficiency Comparison.} }
\label{tab:data_efficiency}
\vspace{-0.3cm}
\end{table}
\medskip
\noindent \textbf{Data Efficiency comparision.}
As shown in Table~\ref{tab:data_efficiency}, LLaVA-OV+ReKV~\cite{di2025rekv} with SOPE training achieves a 2.1-point improvement on OVO-Bench~\cite{liOVOBenchHowFar2025} using only 30K offline video samples, without requiring any stream-specific data. In contrast, StreamForest~\cite{zeng2025streamforestefficientonlinevideo}, relies on orders of magnitude more offline video samples and 121K additional stream-tailored data to attain comparable gains. Moreover, owing to this data efficiency, our method requires only 8 GPUs, merely one quarter of the resources used by StreamForest.
\section{Conclusion}
\label{sec:conclusion}
In this work, we examine how Time-Agnosticism limits the applicability of Video-LLMs in streaming settings, manifesting as two coupled issues: Temporal Order Ambiguity and Past–Current Focus Blindness. We propose WeaveTime, a Video-LLM-agnostic framework that enhances temporal awareness and adaptive memory use. It combines a Temporal Reconstruction auxiliary task to instill streaming order and strengthen causal encoding with a Past–Current Dynamic Focus (PCDF) Cache for uncertainty-aware, coarse-to-fine memory retrieval. Experiments show that WeaveTime significantly improves both accuracy and streaming efficiency over strong baselines.

\clearpage
{
    \small
    \bibliographystyle{ieeenat_fullname}
    \bibliography{main}

@String(ICLR = {Int. Conf. Learn. Represent.})

@String(CVPRW= {IEEE Conf. Comput. Vis. Pattern Recog. Worksh.})

@String(ICLR  = {ICLR})

@String(CVPRW= {CVPRW})

@inproceedings{khattab2020colbert,
  title={Colbert: Efficient and effective passage search via contextualized late interaction over bert},
  author={Khattab, Omar and Zaharia, Matei},
  booktitle={Proceedings of the 43rd International ACM SIGIR conference on research and development in Information Retrieval},
  pages={39--48},
  year={2020}
}

@misc{geminiteam2024gemini15unlockingmultimodal,
      title={Gemini 1.5: Unlocking multimodal understanding across millions of tokens of context}, 
      author={Gemini Team and Petko Georgiev and Ving Ian Lei and Ryan Burnell and Libin Bai and Anmol Gulati and Garrett Tanzer and Damien Vincent and Zhufeng Pan and Shibo Wang and Soroosh Mariooryad and Yifan Ding and others},
      year={2024},
      eprint={2403.05530},
      archivePrefix={arXiv},
      primaryClass={cs.CL},
      url={https://arxiv.org/abs/2403.05530}, 
}

@misc{openai2024gpt4ocard,
      title={GPT-4o System Card}, 
      author={Aaron Hurst and Adam Lerer and Adam P. Goucher and Adam Perelman and Aditya Ramesh and others},
      year={2024},
      eprint={2410.21276},
      archivePrefix={arXiv},
      primaryClass={cs.CL},
      url={https://arxiv.org/abs/2410.21276}, 
}

@misc{huang2024prunevidvisualtokenpruning,
      title={PruneVid: Visual Token Pruning for Efficient Video Large Language Models}, 
      author={Xiaohu Huang and Hao Zhou and Kai Han},
      year={2024},
      eprint={2412.16117},
      archivePrefix={arXiv},
      primaryClass={cs.CV},
      url={https://arxiv.org/abs/2412.16117}, 
}

@misc{pătrăucean2023perceptiontestdiagnosticbenchmark,
      title={Perception Test: A Diagnostic Benchmark for Multimodal Video Models}, 
      author={Viorica Pătrăucean and Lucas Smaira and Ankush Gupta and Adrià Recasens Continente and Larisa Markeeva and Dylan Banarse and Skanda Koppula and Joseph Heyward and Mateusz Malinowski and Yi Yang and Carl Doersch and Tatiana Matejovicova and Yury Sulsky and Antoine Miech and Alex Frechette and Hanna Klimczak and Raphael Koster and Junlin Zhang and Stephanie Winkler and Yusuf Aytar and Simon Osindero and Dima Damen and Andrew Zisserman and João Carreira},
      year={2023},
      eprint={2305.13786},
      archivePrefix={arXiv},
      primaryClass={cs.CV},
      url={https://arxiv.org/abs/2305.13786}, 
}

@misc{li2024mvbenchcomprehensivemultimodalvideo,
      title={MVBench: A Comprehensive Multi-modal Video Understanding Benchmark}, 
      author={Kunchang Li and Yali Wang and Yinan He and Yizhuo Li and Yi Wang and Yi Liu and Zun Wang and Jilan Xu and Guo Chen and Ping Luo and Limin Wang and Yu Qiao},
      year={2024},
      eprint={2311.17005},
      archivePrefix={arXiv},
      primaryClass={cs.CV},
      url={https://arxiv.org/abs/2311.17005}, 
}

@misc{chen2024fargpt4vclosinggap,
      title={How Far Are We to GPT-4V? Closing the Gap to Commercial Multimodal Models with Open-Source Suites}, 
      author={Zhe Chen and Weiyun Wang and Hao Tian and Shenglong Ye and Zhangwei Gao and Erfei Cui and Wenwen Tong and Kongzhi Hu and Jiapeng Luo and Zheng Ma and Ji Ma and Jiaqi Wang and Xiaoyi Dong and Hang Yan and Hewei Guo and Conghui He and Botian Shi and Zhenjiang Jin and Chao Xu and Bin Wang and Xingjian Wei and Wei Li and Wenjian Zhang and Bo Zhang and Pinlong Cai and Licheng Wen and Xiangchao Yan and Min Dou and Lewei Lu and Xizhou Zhu and Tong Lu and Dahua Lin and Yu Qiao and Jifeng Dai and Wenhai Wang},
      year={2024},
      eprint={2404.16821},
      archivePrefix={arXiv},
      primaryClass={cs.CV},
      url={https://arxiv.org/abs/2404.16821}, 
}

@misc{liu2024tempcompassvideollmsreally,
      title={TempCompass: Do Video LLMs Really Understand Videos?}, 
      author={Yuanxin Liu and Shicheng Li and Yi Liu and Yuxiang Wang and Shuhuai Ren and Lei Li and Sishuo Chen and Xu Sun and Lu Hou},
      year={2024},
      eprint={2403.00476},
      archivePrefix={arXiv},
      primaryClass={cs.CV},
      url={https://arxiv.org/abs/2403.00476}, 
}

@misc{zeng2025streamforestefficientonlinevideo,
      title={StreamForest: Efficient Online Video Understanding with Persistent Event Memory}, 
      author={Xiangyu Zeng and Kefan Qiu and Qingyu Zhang and Xinhao Li and Jing Wang and Jiaxin Li and Ziang Yan and Kun Tian and Meng Tian and Xinhai Zhao and Yi Wang and Limin Wang},
      year={2025},
      eprint={2509.24871},
      archivePrefix={arXiv},
      primaryClass={cs.CV},
      url={https://arxiv.org/abs/2509.24871}, 
}

@misc{yang2025streammemqueryagnostickvcache,
      title={StreamMem: Query-Agnostic KV Cache Memory for Streaming Video Understanding}, 
      author={Yanlai Yang and Zhuokai Zhao and Satya Narayan Shukla and Aashu Singh and Shlok Kumar Mishra and Lizhu Zhang and Mengye Ren},
      year={2025},
      eprint={2508.15717},
      archivePrefix={arXiv},
      primaryClass={cs.CV},
      url={https://arxiv.org/abs/2508.15717}, 
}

@misc{liu2023visualinstructiontuning,
      title={Visual Instruction Tuning}, 
      author={Haotian Liu and Chunyuan Li and Qingyang Wu and Yong Jae Lee},
      year={2023},
      eprint={2304.08485},
      archivePrefix={arXiv},
      primaryClass={cs.CV},
      url={https://arxiv.org/abs/2304.08485}, 
}

@misc{kim2025infinipotvmemoryconstrainedkvcache,
      title={InfiniPot-V: Memory-Constrained KV Cache Compression for Streaming Video Understanding}, 
      author={Minsoo Kim and Kyuhong Shim and Jungwook Choi and Simyung Chang},
      year={2025},
      eprint={2506.15745},
      archivePrefix={arXiv},
      primaryClass={eess.IV},
      url={https://arxiv.org/abs/2506.15745}, 
}

@misc{xing2024mitigatingobjecthallucinationconcentric,
      title={Mitigating Object Hallucination via Concentric Causal Attention}, 
      author={Yun Xing and Yiheng Li and Ivan Laptev and Shijian Lu},
      year={2024},
      eprint={2410.15926},
      archivePrefix={arXiv},
      primaryClass={cs.CV},
      url={https://arxiv.org/abs/2410.15926}, 
}

@misc{yang2025svbenchbenchmarktemporalmultiturn,
      title={SVBench: A Benchmark with Temporal Multi-Turn Dialogues for Streaming Video Understanding}, 
      author={Zhenyu Yang and Yuhang Hu and Zemin Du and Dizhan Xue and Shengsheng Qian and Jiahong Wu and Fan Yang and Weiming Dong and Changsheng Xu},
      year={2025},
      eprint={2502.10810},
      archivePrefix={arXiv},
      primaryClass={cs.CV},
      url={https://arxiv.org/abs/2502.10810}, 
}

@misc{huang2025onlinevideounderstandingovbench,
      title={Online Video Understanding: OVBench and VideoChat-Online}, 
      author={Zhenpeng Huang and Xinhao Li and Jiaqi Li and Jing Wang and Xiangyu Zeng and Cheng Liang and Tao Wu and Xi Chen and Liang Li and Limin Wang},
      year={2025},
      eprint={2501.00584},
      archivePrefix={arXiv},
      primaryClass={cs.CV},
      url={https://arxiv.org/abs/2501.00584}, 
}

@misc{zou2025looktwiceanswermemoryspace,
      title={Look Twice Before You Answer: Memory-Space Visual Retracing for Hallucination Mitigation in Multimodal Large Language Models}, 
      author={Xin Zou and Yizhou Wang and Yibo Yan and Yuanhuiyi Lyu and Kening Zheng and Sirui Huang and Junkai Chen and Peijie Jiang and Jia Liu and Chang Tang and Xuming Hu},
      year={2025},
      eprint={2410.03577},
      archivePrefix={arXiv},
      primaryClass={cs.CV},
      url={https://arxiv.org/abs/2410.03577}, 
}

@misc{zhang2025eyeswideopenego,
      title={Eyes Wide Open: Ego Proactive Video-LLM for Streaming Video}, 
      author={Yulin Zhang and Cheng Shi and Yang Wang and Sibei Yang},
      year={2025},
      eprint={2510.14560},
      archivePrefix={arXiv},
      primaryClass={cs.CV},
      url={https://arxiv.org/abs/2510.14560}, 
}

@misc{wang2025streambridgeturningofflinevideo,
      title={StreamBridge: Turning Your Offline Video Large Language Model into a Proactive Streaming Assistant}, 
      author={Haibo Wang and Bo Feng and Zhengfeng Lai and Mingze Xu and Shiyu Li and Weifeng Ge and Afshin Dehghan and Meng Cao and Ping Huang},
      year={2025},
      eprint={2505.05467},
      archivePrefix={arXiv},
      primaryClass={cs.CV},
      url={https://arxiv.org/abs/2505.05467}, 
}

@inproceedings{di2025rekv,
  title={Streaming Video Question-Answering with In-context Video KV-Cache Retrieval},
  author={Di, Shangzhe and Yu, Zhelun and Zhang, Guanghao and Li, Haoyuan and Cheng, Hao and Li, Bolin and He, Wanggui and Shu, Fangxun and Jiang, Hao and others},
  booktitle={ICLR},
  year={2025}
}

@misc{zhai2023sigmoidlosslanguageimage,
      title={Sigmoid Loss for Language Image Pre-Training}, 
      author={Xiaohua Zhai and Basil Mustafa and Alexander Kolesnikov and Lucas Beyer},
      year={2023},
      eprint={2303.15343},
      archivePrefix={arXiv},
      primaryClass={cs.CV},
      url={https://arxiv.org/abs/2303.15343}, 
}

@misc{ataallah2024infinibenchcomprehensivebenchmarklarge,
      title={InfiniBench: A Benchmark for Large Multi-Modal Models in Long-Form Movies and TV Shows}, 
      author={Kirolos Ataallah and Eslam Abdelrahman and Mahmoud Ahmed and Chenhui Gou and Khushbu Pahwa and Jian Ding and Mohamed Elhoseiny},
      year={2025},
      eprint={2406.19875},
      archivePrefix={arXiv},
      primaryClass={cs.CV},
      url={https://arxiv.org/abs/2406.19875}, 
}

@misc{wang2024lvbenchextremelongvideo,
      title={LVBench: An Extreme Long Video Understanding Benchmark}, 
      author={Weihan Wang and Zehai He and Wenyi Hong and Yean Cheng and Xiaohan Zhang and Ji Qi and Xiaotao Gu and Shiyu Huang and Bin Xu and Yuxiao Dong and Ming Ding and Jie Tang},
      year={2025},
      eprint={2406.08035},
      archivePrefix={arXiv},
      primaryClass={cs.CV},
      url={https://arxiv.org/abs/2406.08035}, 
}

@misc{grattafiori2024llama3herdmodels,
      title={The Llama 3 Herd of Models}, 
      author={Aaron Grattafiori and Abhimanyu Dubey and Abhinav Jauhri and Abhinav Pandey and Abhishek Kadian and Ahmad Al-Dahle and Aiesha Letman and Akhil Mathur and Alan Schelten and Alex Vaughan and Amy Yang and Angela Fan and Anirudh Goyal and Anthony Hartshorn and Aobo Yang and Archi Mitra and Archie Sravankumar and Artem Korenev and Arthur Hinsvark and Arun Rao and Aston Zhang and Aurelien Rodriguez and Austen Gregerson and Ava Spataru and Baptiste Roziere and Bethany Biron and Binh Tang and Bobbie Chern and Charlotte Caucheteux and Chaya Nayak and Chloe Bi and Chris Marra and Chris McConnell and Christian Keller and Christophe Touret and Chunyang Wu and Corinne Wong and Cristian Canton Ferrer and Cyrus Nikolaidis and Damien Allonsius and Daniel Song and Danielle Pintz and Danny Livshits and Danny Wyatt and David Esiobu and Dhruv Choudhary and Dhruv Mahajan and Diego Garcia-Olano and Diego Perino and Dieuwke Hupkes and Egor Lakomkin and Ehab AlBadawy and Elina Lobanova and Emily Dinan and Eric Michael Smith and Filip Radenovic and Francisco Guzmán and Frank Zhang and Gabriel Synnaeve and Gabrielle Lee and Georgia Lewis Anderson and Govind Thattai and Graeme Nail and Gregoire Mialon and Guan Pang and Guillem Cucurell and Hailey Nguyen and Hannah Korevaar and Hu Xu and Hugo Touvron and Iliyan Zarov and Imanol Arrieta Ibarra and Isabel Kloumann and Ishan Misra and Ivan Evtimov and Jack Zhang and Jade Copet and Jaewon Lee and Jan Geffert and Jana Vranes and Jason Park and Jay Mahadeokar and Jeet Shah and Jelmer van der Linde and Jennifer Billock and Jenny Hong and Jenya Lee and Jeremy Fu and Jianfeng Chi and Jianyu Huang and Jiawen Liu and Jie Wang and Jiecao Yu and Joanna Bitton and Joe Spisak and Jongsoo Park and Joseph Rocca and Joshua Johnstun and Joshua Saxe and Junteng Jia and Kalyan Vasuden Alwala and Karthik Prasad and Kartikeya Upasani and Kate Plawiak and Ke Li and Kenneth Heafield and Kevin Stone and Khalid El-Arini and Krithika Iyer and Kshitiz Malik and Kuenley Chiu and Kunal Bhalla and Kushal Lakhotia and Lauren Rantala-Yeary and Laurens van der Maaten and Lawrence Chen and Liang Tan and Liz Jenkins and Louis Martin and Lovish Madaan and Lubo Malo and Lukas Blecher and Lukas Landzaat and Luke de Oliveira and Madeline Muzzi and Mahesh Pasupuleti and Mannat Singh and Manohar Paluri and Marcin Kardas and Maria Tsimpoukelli and Mathew Oldham and Mathieu Rita and Maya Pavlova and Melanie Kambadur and Mike Lewis and Min Si and Mitesh Kumar Singh and Mona Hassan and Naman Goyal and Narjes Torabi and Nikolay Bashlykov and Nikolay Bogoychev and Niladri Chatterji and Ning Zhang and Olivier Duchenne and Onur Çelebi and Patrick Alrassy and Pengchuan Zhang and Pengwei Li and Petar Vasic and Peter Weng and Prajjwal Bhargava and Pratik Dubal and Praveen Krishnan and Punit Singh Koura and Puxin Xu and Qing He and Qingxiao Dong and Ragavan Srinivasan and Raj Ganapathy and Ramon Calderer and Ricardo Silveira Cabral and Robert Stojnic and Roberta Raileanu and Rohan Maheswari and Rohit Girdhar and Rohit Patel and Romain Sauvestre and Ronnie Polidoro and Roshan Sumbaly and Ross Taylor and Ruan Silva and Rui Hou and Rui Wang and Saghar Hosseini and Sahana Chennabasappa and Sanjay Singh and Sean Bell and Seohyun Sonia Kim and Sergey Edunov and Shaoliang Nie and Sharan Narang and Sharath Raparthy and Sheng Shen and Shengye Wan and Shruti Bhosale and Shun Zhang and Simon Vandenhende and Soumya Batra and Spencer Whitman and Sten Sootla and Stephane Collot and Suchin Gururangan and Sydney Borodinsky and Tamar Herman and Tara Fowler and Tarek Sheasha and Thomas Georgiou and Thomas Scialom and Tobias Speckbacher and Todor Mihaylov and Tong Xiao and Ujjwal Karn and Vedanuj Goswami and Vibhor Gupta and Vignesh Ramanathan and Viktor Kerkez and Vincent Gonguet and Virginie Do and Vish Vogeti and Vítor Albiero and Vladan Petrovic and Weiwei Chu and Wenhan Xiong and Wenyin Fu and Whitney Meers and Xavier Martinet and Xiaodong Wang and Xiaofang Wang and Xiaoqing Ellen Tan and Xide Xia and Xinfeng Xie and Xuchao Jia and Xuewei Wang and Yaelle Goldschlag and Yashesh Gaur and Yasmine Babaei and Yi Wen and Yiwen Song and Yuchen Zhang and Yue Li and Yuning Mao and Zacharie Delpierre Coudert and Zheng Yan and Zhengxing Chen and Zoe Papakipos and Aaditya Singh and Aayushi Srivastava and Abha Jain and Adam Kelsey and Adam Shajnfeld and Adithya Gangidi and Adolfo Victoria and Ahuva Goldstand and Ajay Menon and Ajay Sharma and Alex Boesenberg and Alexei Baevski and Allie Feinstein and Amanda Kallet and Amit Sangani and Amos Teo and Anam Yunus and Andrei Lupu and Andres Alvarado and Andrew Caples and Andrew Gu and Andrew Ho and Andrew Poulton and Andrew Ryan and Ankit Ramchandani and Annie Dong and Annie Franco and Anuj Goyal and Aparajita Saraf and Arkabandhu Chowdhury and Ashley Gabriel and Ashwin Bharambe and Assaf Eisenman and Azadeh Yazdan and Beau James and Ben Maurer and Benjamin Leonhardi and Bernie Huang and Beth Loyd and Beto De Paola and Bhargavi Paranjape and Bing Liu and Bo Wu and Boyu Ni and Braden Hancock and Bram Wasti and Brandon Spence and Brani Stojkovic and Brian Gamido and Britt Montalvo and Carl Parker and Carly Burton and Catalina Mejia and Ce Liu and Changhan Wang and Changkyu Kim and Chao Zhou and Chester Hu and Ching-Hsiang Chu and Chris Cai and Chris Tindal and Christoph Feichtenhofer and Cynthia Gao and Damon Civin and Dana Beaty and Daniel Kreymer and Daniel Li and David Adkins and David Xu and Davide Testuggine and Delia David and Devi Parikh and Diana Liskovich and Didem Foss and Dingkang Wang and Duc Le and Dustin Holland and Edward Dowling and Eissa Jamil and Elaine Montgomery and Eleonora Presani and Emily Hahn and Emily Wood and Eric-Tuan Le and Erik Brinkman and Esteban Arcaute and Evan Dunbar and Evan Smothers and Fei Sun and Felix Kreuk and Feng Tian and Filippos Kokkinos and Firat Ozgenel and Francesco Caggioni and Frank Kanayet and Frank Seide and Gabriela Medina Florez and Gabriella Schwarz and Gada Badeer and Georgia Swee and Gil Halpern and Grant Herman and Grigory Sizov and Guangyi and Zhang and Guna Lakshminarayanan and Hakan Inan and Hamid Shojanazeri and Han Zou and Hannah Wang and Hanwen Zha and Haroun Habeeb and Harrison Rudolph and Helen Suk and Henry Aspegren and Hunter Goldman and Hongyuan Zhan and Ibrahim Damlaj and Igor Molybog and Igor Tufanov and Ilias Leontiadis and Irina-Elena Veliche and Itai Gat and Jake Weissman and James Geboski and James Kohli and Janice Lam and Japhet Asher and Jean-Baptiste Gaya and Jeff Marcus and Jeff Tang and Jennifer Chan and Jenny Zhen and Jeremy Reizenstein and Jeremy Teboul and Jessica Zhong and Jian Jin and Jingyi Yang and Joe Cummings and Jon Carvill and Jon Shepard and Jonathan McPhie and Jonathan Torres and Josh Ginsburg and Junjie Wang and Kai Wu and Kam Hou U and Karan Saxena and Kartikay Khandelwal and Katayoun Zand and Kathy Matosich and Kaushik Veeraraghavan and Kelly Michelena and Keqian Li and Kiran Jagadeesh and Kun Huang and Kunal Chawla and Kyle Huang and Lailin Chen and Lakshya Garg and Lavender A and Leandro Silva and Lee Bell and Lei Zhang and Liangpeng Guo and Licheng Yu and Liron Moshkovich and Luca Wehrstedt and Madian Khabsa and Manav Avalani and Manish Bhatt and Martynas Mankus and Matan Hasson and Matthew Lennie and Matthias Reso and Maxim Groshev and Maxim Naumov and Maya Lathi and Meghan Keneally and Miao Liu and Michael L. Seltzer and Michal Valko and Michelle Restrepo and Mihir Patel and Mik Vyatskov and Mikayel Samvelyan and Mike Clark and Mike Macey and Mike Wang and Miquel Jubert Hermoso and Mo Metanat and Mohammad Rastegari and Munish Bansal and Nandhini Santhanam and Natascha Parks and Natasha White and Navyata Bawa and Nayan Singhal and Nick Egebo and Nicolas Usunier and Nikhil Mehta and Nikolay Pavlovich Laptev and Ning Dong and Norman Cheng and Oleg Chernoguz and Olivia Hart and Omkar Salpekar and Ozlem Kalinli and Parkin Kent and Parth Parekh and Paul Saab and Pavan Balaji and Pedro Rittner and Philip Bontrager and Pierre Roux and Piotr Dollar and Polina Zvyagina and Prashant Ratanchandani and Pritish Yuvraj and Qian Liang and Rachad Alao and Rachel Rodriguez and Rafi Ayub and Raghotham Murthy and Raghu Nayani and Rahul Mitra and Rangaprabhu Parthasarathy and Raymond Li and Rebekkah Hogan and Robin Battey and Rocky Wang and Russ Howes and Ruty Rinott and Sachin Mehta and Sachin Siby and Sai Jayesh Bondu and Samyak Datta and Sara Chugh and Sara Hunt and Sargun Dhillon and Sasha Sidorov and Satadru Pan and Saurabh Mahajan and Saurabh Verma and Seiji Yamamoto and Sharadh Ramaswamy and Shaun Lindsay and Shaun Lindsay and Sheng Feng and Shenghao Lin and Shengxin Cindy Zha and Shishir Patil and Shiva Shankar and Shuqiang Zhang and Shuqiang Zhang and Sinong Wang and Sneha Agarwal and Soji Sajuyigbe and Soumith Chintala and Stephanie Max and Stephen Chen and Steve Kehoe and Steve Satterfield and Sudarshan Govindaprasad and Sumit Gupta and Summer Deng and Sungmin Cho and Sunny Virk and Suraj Subramanian and Sy Choudhury and Sydney Goldman and Tal Remez and Tamar Glaser and Tamara Best and Thilo Koehler and Thomas Robinson and Tianhe Li and Tianjun Zhang and Tim Matthews and Timothy Chou and Tzook Shaked and Varun Vontimitta and Victoria Ajayi and Victoria Montanez and Vijai Mohan and Vinay Satish Kumar and Vishal Mangla and Vlad Ionescu and Vlad Poenaru and Vlad Tiberiu Mihailescu and Vladimir Ivanov and Wei Li and Wenchen Wang and Wenwen Jiang and Wes Bouaziz and Will Constable and Xiaocheng Tang and Xiaojian Wu and Xiaolan Wang and Xilun Wu and Xinbo Gao and Yaniv Kleinman and Yanjun Chen and Ye Hu and Ye Jia and Ye Qi and Yenda Li and Yilin Zhang and Ying Zhang and Yossi Adi and Youngjin Nam and Yu and Wang and Yu Zhao and Yuchen Hao and Yundi Qian and Yunlu Li and Yuzi He and Zach Rait and Zachary DeVito and Zef Rosnbrick and Zhaoduo Wen and Zhenyu Yang and Zhiwei Zhao and Zhiyu Ma},
      year={2024},
      eprint={2407.21783},
      archivePrefix={arXiv},
      primaryClass={cs.AI},
      url={https://arxiv.org/abs/2407.21783}, 
}

@misc{deepseekai2025deepseekv3technicalreport,
      title={DeepSeek-V3 Technical Report}, 
      author={DeepSeek-AI and Aixin Liu and Bei Feng and Bing Xue and Bingxuan Wang and Bochao Wu and Chengda Lu and Chenggang Zhao and Chengqi Deng and Chenyu Zhang and Chong Ruan and Damai Dai and Daya Guo and Dejian Yang and Deli Chen and Dongjie Ji and Erhang Li and Fangyun Lin and Fucong Dai and Fuli Luo and Guangbo Hao and Guanting Chen and Guowei Li and H. Zhang and Han Bao and Hanwei Xu and Haocheng Wang and Haowei Zhang and Honghui Ding and Huajian Xin and Huazuo Gao and Hui Li and Hui Qu and J. L. Cai and Jian Liang and Jianzhong Guo and Jiaqi Ni and Jiashi Li and Jiawei Wang and Jin Chen and Jingchang Chen and Jingyang Yuan and Junjie Qiu and Junlong Li and Junxiao Song and Kai Dong and Kai Hu and Kaige Gao and Kang Guan and Kexin Huang and Kuai Yu and Lean Wang and Lecong Zhang and Lei Xu and Leyi Xia and Liang Zhao and Litong Wang and Liyue Zhang and Meng Li and Miaojun Wang and Mingchuan Zhang and Minghua Zhang and Minghui Tang and Mingming Li and Ning Tian and Panpan Huang and Peiyi Wang and Peng Zhang and Qiancheng Wang and Qihao Zhu and Qinyu Chen and Qiushi Du and R. J. Chen and R. L. Jin and Ruiqi Ge and Ruisong Zhang and Ruizhe Pan and Runji Wang and Runxin Xu and Ruoyu Zhang and Ruyi Chen and S. S. Li and Shanghao Lu and Shangyan Zhou and Shanhuang Chen and Shaoqing Wu and Shengfeng Ye and Shengfeng Ye and Shirong Ma and Shiyu Wang and Shuang Zhou and Shuiping Yu and Shunfeng Zhou and Shuting Pan and T. Wang and Tao Yun and Tian Pei and Tianyu Sun and W. L. Xiao and Wangding Zeng and Wanjia Zhao and Wei An and Wen Liu and Wenfeng Liang and Wenjun Gao and Wenqin Yu and Wentao Zhang and X. Q. Li and Xiangyue Jin and Xianzu Wang and Xiao Bi and Xiaodong Liu and Xiaohan Wang and Xiaojin Shen and Xiaokang Chen and Xiaokang Zhang and Xiaosha Chen and Xiaotao Nie and Xiaowen Sun and Xiaoxiang Wang and Xin Cheng and Xin Liu and Xin Xie and Xingchao Liu and Xingkai Yu and Xinnan Song and Xinxia Shan and Xinyi Zhou and Xinyu Yang and Xinyuan Li and Xuecheng Su and Xuheng Lin and Y. K. Li and Y. Q. Wang and Y. X. Wei and Y. X. Zhu and Yang Zhang and Yanhong Xu and Yanhong Xu and Yanping Huang and Yao Li and Yao Zhao and Yaofeng Sun and Yaohui Li and Yaohui Wang and Yi Yu and Yi Zheng and Yichao Zhang and Yifan Shi and Yiliang Xiong and Ying He and Ying Tang and Yishi Piao and Yisong Wang and Yixuan Tan and Yiyang Ma and Yiyuan Liu and Yongqiang Guo and Yu Wu and Yuan Ou and Yuchen Zhu and Yuduan Wang and Yue Gong and Yuheng Zou and Yujia He and Yukun Zha and Yunfan Xiong and Yunxian Ma and Yuting Yan and Yuxiang Luo and Yuxiang You and Yuxuan Liu and Yuyang Zhou and Z. F. Wu and Z. Z. Ren and Zehui Ren and Zhangli Sha and Zhe Fu and Zhean Xu and Zhen Huang and Zhen Zhang and Zhenda Xie and Zhengyan Zhang and Zhewen Hao and Zhibin Gou and Zhicheng Ma and Zhigang Yan and Zhihong Shao and Zhipeng Xu and Zhiyu Wu and Zhongyu Zhang and Zhuoshu Li and Zihui Gu and Zijia Zhu and Zijun Liu and Zilin Li and Ziwei Xie and Ziyang Song and Ziyi Gao and Zizheng Pan},
      year={2025},
      eprint={2412.19437},
      archivePrefix={arXiv},
      primaryClass={cs.CL},
      url={https://arxiv.org/abs/2412.19437}, 
}

@misc{wang2024qwen2vlenhancingvisionlanguagemodels,
      title={Qwen2-VL: Enhancing Vision-Language Model's Perception of the World at Any Resolution}, 
      author={Peng Wang and Shuai Bai and Sinan Tan and Shijie Wang and Zhihao Fan and Jinze Bai and Keqin Chen and Xuejing Liu and Jialin Wang and Wenbin Ge and Yang Fan and Kai Dang and Mengfei Du and Xuancheng Ren and Rui Men and Dayiheng Liu and Chang Zhou and Jingren Zhou and Junyang Lin},
      year={2024},
      eprint={2409.12191},
      archivePrefix={arXiv},
      primaryClass={cs.CV},
      url={https://arxiv.org/abs/2409.12191}, 
}

@misc{li2024llavanextinterleavetacklingmultiimagevideo,
      title={LLaVA-NeXT-Interleave: Tackling Multi-image, Video, and 3D in Large Multimodal Models}, 
      author={Feng Li and Renrui Zhang and Hao Zhang and Yuanhan Zhang and Bo Li and Wei Li and Zejun Ma and Chunyuan Li},
      year={2024},
      eprint={2407.07895},
      archivePrefix={arXiv},
      primaryClass={cs.CV},
      url={https://arxiv.org/abs/2407.07895}, 
}

@misc{li2024llavaonevisioneasyvisualtask,
      title={LLaVA-OneVision: Easy Visual Task Transfer}, 
      author={Bo Li and Yuanhan Zhang and Dong Guo and Renrui Zhang and Feng Li and Hao Zhang and Kaichen Zhang and Peiyuan Zhang and Yanwei Li and Ziwei Liu and Chunyuan Li},
      year={2024},
      eprint={2408.03326},
      archivePrefix={arXiv},
      primaryClass={cs.CV},
      url={https://arxiv.org/abs/2408.03326}, 
}

@misc{yao2024minicpmvgpt4vlevelmllm,
      title={MiniCPM-V: A GPT-4V Level MLLM on Your Phone}, 
      author={Yuan Yao and Tianyu Yu and Ao Zhang and Chongyi Wang and Junbo Cui and Hongji Zhu and Tianchi Cai and Haoyu Li and Weilin Zhao and Zhihui He and Qianyu Chen and Huarong Zhou and Zhensheng Zou and Haoye Zhang and Shengding Hu and Zhi Zheng and Jie Zhou and Jie Cai and Xu Han and Guoyang Zeng and Dahai Li and Zhiyuan Liu and Maosong Sun},
      year={2024},
      eprint={2408.01800},
      archivePrefix={arXiv},
      primaryClass={cs.CV},
      url={https://arxiv.org/abs/2408.01800}, 
}

@misc{radford2021learningtransferablevisualmodels,
      title={Learning Transferable Visual Models From Natural Language Supervision}, 
      author={Alec Radford and Jong Wook Kim and Chris Hallacy and Aditya Ramesh and Gabriel Goh and Sandhini Agarwal and Girish Sastry and Amanda Askell and Pamela Mishkin and Jack Clark and Gretchen Krueger and Ilya Sutskever},
      year={2021},
      eprint={2103.00020},
      archivePrefix={arXiv},
      primaryClass={cs.CV},
      url={https://arxiv.org/abs/2103.00020}, 
}

@misc{vaswani2023attentionneed,
      title={Attention Is All You Need}, 
      author={Ashish Vaswani and Noam Shazeer and Niki Parmar and Jakob Uszkoreit and Llion Jones and Aidan N. Gomez and Lukasz Kaiser and Illia Polosukhin},
      year={2023},
      eprint={1706.03762},
      archivePrefix={arXiv},
      primaryClass={cs.CL},
      url={https://arxiv.org/abs/1706.03762}, 
}

@misc{baiSequentialModelingEnables2023,
      title={Sequential Modeling Enables Scalable Learning for Large Vision Models}, 
      author={Yutong Bai and Xinyang Geng and Karttikeya Mangalam and Amir Bar and Alan Yuille and Trevor Darrell and Jitendra Malik and Alexei A Efros},
      year={2023},
      eprint={2312.00785},
      archivePrefix={arXiv},
      primaryClass={cs.CV},
      url={https://arxiv.org/abs/2312.00785}, 
}

@misc{sharmaLosingVisualNeedles2024,
      title={Losing Visual Needles in Image Haystacks: Vision Language Models are Easily Distracted in Short and Long Contexts}, 
      author={Aditya Sharma and Michael Saxon and William Yang Wang},
      year={2024},
      eprint={2406.16851},
      archivePrefix={arXiv},
      primaryClass={cs.CL},
      url={https://arxiv.org/abs/2406.16851}, 
}

@misc{mangalamEgoSchemaDiagnosticBenchmark2023,
      title={EgoSchema: A Diagnostic Benchmark for Very Long-form Video Language Understanding}, 
      author={Karttikeya Mangalam and Raiymbek Akshulakov and Jitendra Malik},
      year={2023},
      eprint={2308.09126},
      archivePrefix={arXiv},
      primaryClass={cs.CV},
      url={https://arxiv.org/abs/2308.09126}, 
}

@misc{graumanEgo4DWorld30002022,
      title={Ego4D: Around the World in 3,000 Hours of Egocentric Video}, 
      author={Kristen Grauman and Andrew Westbury and Eugene Byrne and Zachary Chavis and Antonino Furnari and Rohit Girdhar and Jackson Hamburger and Hao Jiang and Miao Liu and Xingyu Liu and Miguel Martin and Tushar Nagarajan and Ilija Radosavovic and Santhosh Kumar Ramakrishnan and Fiona Ryan and Jayant Sharma and Michael Wray and Mengmeng Xu and Eric Zhongcong Xu and Chen Zhao and Siddhant Bansal and Dhruv Batra and Vincent Cartillier and Sean Crane and Tien Do and Morrie Doulaty and Akshay Erapalli and Christoph Feichtenhofer and Adriano Fragomeni and Qichen Fu and Abrham Gebreselasie and Cristina Gonzalez and James Hillis and Xuhua Huang and Yifei Huang and Wenqi Jia and Weslie Khoo and Jachym Kolar and Satwik Kottur and Anurag Kumar and Federico Landini and Chao Li and Yanghao Li and Zhenqiang Li and Karttikeya Mangalam and Raghava Modhugu and Jonathan Munro and Tullie Murrell and Takumi Nishiyasu and Will Price and Paola Ruiz Puentes and Merey Ramazanova and Leda Sari and Kiran Somasundaram and Audrey Southerland and Yusuke Sugano and Ruijie Tao and Minh Vo and Yuchen Wang and Xindi Wu and Takuma Yagi and Ziwei Zhao and Yunyi Zhu and Pablo Arbelaez and David Crandall and Dima Damen and Giovanni Maria Farinella and Christian Fuegen and Bernard Ghanem and Vamsi Krishna Ithapu and C. V. Jawahar and Hanbyul Joo and Kris Kitani and Haizhou Li and Richard Newcombe and Aude Oliva and Hyun Soo Park and James M. Rehg and Yoichi Sato and Jianbo Shi and Mike Zheng Shou and Antonio Torralba and Lorenzo Torresani and Mingfei Yan and Jitendra Malik},
      year={2022},
      eprint={2110.07058},
      archivePrefix={arXiv},
      primaryClass={cs.CV},
      url={https://arxiv.org/abs/2110.07058}, 
}

@misc{wuVideoLLMMoDEfficientVideoLanguage2024,
      title={VideoLLM-MoD: Efficient Video-Language Streaming with Mixture-of-Depths Vision Computation}, 
      author={Shiwei Wu and Joya Chen and Kevin Qinghong Lin and Qimeng Wang and Yan Gao and Qianli Xu and Tong Xu and Yao Hu and Enhong Chen and Mike Zheng Shou},
      year={2024},
      eprint={2408.16730},
      archivePrefix={arXiv},
      primaryClass={cs.CV},
      url={https://arxiv.org/abs/2408.16730}, 
}

@misc{chenVideoLLMonlineOnlineVideo2024,
      title={VideoLLM-online: Online Video Large Language Model for Streaming Video}, 
      author={Joya Chen and Zhaoyang Lv and Shiwei Wu and Kevin Qinghong Lin and Chenan Song and Difei Gao and Jia-Wei Liu and Ziteng Gao and Dongxing Mao and Mike Zheng Shou},
      year={2024},
      eprint={2406.11816},
      archivePrefix={arXiv},
      primaryClass={cs.CV},
      url={https://arxiv.org/abs/2406.11816}, 
}

@misc{yeMMEgoBuildingEgocentric2024,
      title={MM-Ego: Towards Building Egocentric Multimodal LLMs for Video QA}, 
      author={Hanrong Ye and Haotian Zhang and Erik Daxberger and Lin Chen and Zongyu Lin and Yanghao Li and Bowen Zhang and Haoxuan You and Dan Xu and Zhe Gan and Jiasen Lu and Yinfei Yang},
      year={2025},
      eprint={2410.07177},
      archivePrefix={arXiv},
      primaryClass={cs.CV},
      url={https://arxiv.org/abs/2410.07177}, 
}

@misc{shuVideoXLExtraLongVision2024,
      title={Video-XL: Extra-Long Vision Language Model for Hour-Scale Video Understanding}, 
      author={Yan Shu and Zheng Liu and Peitian Zhang and Minghao Qin and Junjie Zhou and Zhengyang Liang and Tiejun Huang and Bo Zhao},
      year={2024},
      eprint={2409.14485},
      archivePrefix={arXiv},
      primaryClass={cs.CV},
      url={https://arxiv.org/abs/2409.14485}, 
}

@misc{kimOpenVLAOpenSourceVisionLanguageAction2024,
      title={OpenVLA: An Open-Source Vision-Language-Action Model}, 
      author={Moo Jin Kim and Karl Pertsch and Siddharth Karamcheti and Ted Xiao and Ashwin Balakrishna and Suraj Nair and Rafael Rafailov and Ethan Foster and Grace Lam and Pannag Sanketi and Quan Vuong and Thomas Kollar and Benjamin Burchfiel and Russ Tedrake and Dorsa Sadigh and Sergey Levine and Percy Liang and Chelsea Finn},
      year={2024},
      eprint={2406.09246},
      archivePrefix={arXiv},
      primaryClass={cs.RO},
      url={https://arxiv.org/abs/2406.09246}, 
}

@misc{qianStreamingLongVideo2024,
      title={Streaming Long Video Understanding with Large Language Models}, 
      author={Rui Qian and Xiaoyi Dong and Pan Zhang and Yuhang Zang and Shuangrui Ding and Dahua Lin and Jiaqi Wang},
      year={2024},
      eprint={2405.16009},
      archivePrefix={arXiv},
      primaryClass={cs.CV},
      url={https://arxiv.org/abs/2405.16009}, 
}

@misc{florenceImplicitBehavioralCloning2021,
      title={Implicit Behavioral Cloning}, 
      author={Pete Florence and Corey Lynch and Andy Zeng and Oscar Ramirez and Ayzaan Wahid and Laura Downs and Adrian Wong and Johnny Lee and Igor Mordatch and Jonathan Tompson},
      year={2021},
      eprint={2109.00137},
      archivePrefix={arXiv},
      primaryClass={cs.RO},
      url={https://arxiv.org/abs/2109.00137}, 
}

@misc{zhangFlashVStreamMemoryBasedRealTime2024,
      title={Flash-VStream: Memory-Based Real-Time Understanding for Long Video Streams}, 
      author={Haoji Zhang and Yiqin Wang and Yansong Tang and Yong Liu and Jiashi Feng and Jifeng Dai and Xiaojie Jin},
      year={2024},
      eprint={2406.08085},
      archivePrefix={arXiv},
      primaryClass={cs.CV},
      url={https://arxiv.org/abs/2406.08085}, 
}

@inproceedings{barmannWhereDidLeave2022,
	title = {Where did I leave my keys? — Episodic-Memory-Based Question Answering on Egocentric Videos},
	url = {https://ieeexplore.ieee.org/document/9857465/?arnumber=9857465},
	doi = {10.1109/CVPRW56347.2022.00162},
	shorttitle = {Where did I leave my keys?},
	eventtitle = {2022 {IEEE}/{CVF} Conference on Computer Vision and Pattern Recognition Workshops ({CVPRW})},
year={2024},
	pages = {1559--1567},
	booktitle = {2022 {IEEE}/{CVF} Conference on Computer Vision and Pattern Recognition Workshops ({CVPRW})},
	author = {Bärmann, Leonard and Waibel, Alex},
	urldate = {2024-12-08},
	date = {2022-06},
	note = {{ISSN}: 2160-7516},
}

@misc{taoDyCokeDynamicCompression2024,
      title={DyCoke: Dynamic Compression of Tokens for Fast Video Large Language Models}, 
      author={Keda Tao and Can Qin and Haoxuan You and Yang Sui and Huan Wang},
      year={2025},
      eprint={2411.15024},
      archivePrefix={arXiv},
      primaryClass={cs.CV},
      url={https://arxiv.org/abs/2411.15024}, 
}

@misc{wang2024videollmknowsspeakenhancing,
      title={VideoLLM Knows When to Speak: Enhancing Time-Sensitive Video Comprehension with Video-Text Duet Interaction Format}, 
      author={Yueqian Wang and Xiaojun Meng and Yuxuan Wang and Jianxin Liang and Jiansheng Wei and Huishuai Zhang and Dongyan Zhao},
      year={2024},
      eprint={2411.17991},
      archivePrefix={arXiv},
      primaryClass={cs.CV},
      url={https://arxiv.org/abs/2411.17991}, 
}

@inproceedings{linStreamingBenchAssessingGap2024,
	title = {{StreamingBench}: Assessing the Gap for {MLLMs} to Achieve Streaming Video Understanding},
	url = {https://www.semanticscholar.org/paper/958e80006e3b62a0b41e50877d9e554c339615a7},
    booktitle={None},
	shorttitle = {{StreamingBench}},
	author = {Lin, Junming and Fang, Zheng and Chen, Chi and Wan, Zihao and Luo, Fuwen and Li, Peng and Liu, Yang and Sun, Maosong},
	urldate = {2024-12-13},
year={2024},
	date = {2024-11-06},
}

@misc{zhengDoe1ClosedLoopAutonomous2024,
      title={Doe-1: Closed-Loop Autonomous Driving with Large World Model}, 
      author={Wenzhao Zheng and Zetian Xia and Yuanhui Huang and Sicheng Zuo and Jie Zhou and Jiwen Lu},
      year={2024},
      eprint={2412.09627},
      archivePrefix={arXiv},
      primaryClass={cs.CV},
      url={https://arxiv.org/abs/2412.09627}, 
}

@misc{wangOmniDriveHolisticLLMAgent2024,
      title={OmniDrive: A Holistic Vision-Language Dataset for Autonomous Driving with Counterfactual Reasoning}, 
      author={Shihao Wang and Zhiding Yu and Xiaohui Jiang and Shiyi Lan and Min Shi and Nadine Chang and Jan Kautz and Ying Li and Jose M. Alvarez},
      year={2025},
      eprint={2405.01533},
      archivePrefix={arXiv},
      primaryClass={cs.CV},
      url={https://arxiv.org/abs/2405.01533}, 
}

@misc{zhouMLVUComprehensiveBenchmark2024,
      title={MLVU: Benchmarking Multi-task Long Video Understanding}, 
      author={Junjie Zhou and Yan Shu and Bo Zhao and Boya Wu and Zhengyang Liang and Shitao Xiao and Minghao Qin and Xi Yang and Yongping Xiong and Bo Zhang and Tiejun Huang and Zheng Liu},
      year={2025},
      eprint={2406.04264},
      archivePrefix={arXiv},
      primaryClass={cs.CV},
      url={https://arxiv.org/abs/2406.04264}, 
}

@misc{fuVideoMMEFirstEverComprehensive2024,
      title={Video-MME: The First-Ever Comprehensive Evaluation Benchmark of Multi-modal LLMs in Video Analysis}, 
      author={Chaoyou Fu and Yuhan Dai and Yongdong Luo and Lei Li and Shuhuai Ren and Renrui Zhang and Zihan Wang and others},
      year={2025},
      eprint={2405.21075},
      archivePrefix={arXiv},
      primaryClass={cs.CV},
      url={https://arxiv.org/abs/2405.21075}, 
}

@misc{huangOnlineVideoUnderstanding2024,
      title={Online Video Understanding: OVBench and VideoChat-Online}, 
      author={Zhenpeng Huang and Xinhao Li and Jiaqi Li and Jing Wang and Xiangyu Zeng and Cheng Liang and Tao Wu and Xi Chen and Liang Li and Limin Wang},
      year={2025},
      eprint={2501.00584},
      archivePrefix={arXiv},
      primaryClass={cs.CV},
      url={https://arxiv.org/abs/2501.00584}, 
}

@misc{qianDispiderEnablingVideo2025,
      title={Dispider: Enabling Video LLMs with Active Real-Time Interaction via Disentangled Perception, Decision, and Reaction}, 
      author={Rui Qian and Shuangrui Ding and Xiaoyi Dong and Pan Zhang and Yuhang Zang and Yuhang Cao and Dahua Lin and Jiaqi Wang},
      year={2025},
      eprint={2501.03218},
      archivePrefix={arXiv},
      primaryClass={cs.CV},
      url={https://arxiv.org/abs/2501.03218}, 
}

@misc{liOVOBenchHowFar2025,
      title={OVO-Bench: How Far is Your Video-LLMs from Real-World Online Video Understanding?}, 
      author={Yifei Li and Junbo Niu and Ziyang Miao and Chunjiang Ge and Yuanhang Zhou and Qihao He and Xiaoyi Dong and Haodong Duan and Shuangrui Ding and Rui Qian and Pan Zhang and Yuhang Zang and Yuhang Cao and Conghui He and Jiaqi Wang},
      year={2025},
      eprint={2501.05510},
      archivePrefix={arXiv},
      primaryClass={cs.CV},
      url={https://arxiv.org/abs/2501.05510}, 
}

@misc{liuVideoXLProReconstructiveToken2025,
      title={Video-XL-Pro: Reconstructive Token Compression for Extremely Long Video Understanding}, 
      author={Xiangrui Liu and Yan Shu and Zheng Liu and Ao Li and Yang Tian and Bo Zhao},
      year={2025},
      eprint={2503.18478},
      archivePrefix={arXiv},
      primaryClass={cs.CV},
      url={https://arxiv.org/abs/2503.18478}, 
}

@misc{deng3DLLaVAGeneralist3D2025,
      title={3D-LLaVA: Towards Generalist 3D LMMs with Omni Superpoint Transformer}, 
      author={Jiajun Deng and Tianyu He and Li Jiang and Tianyu Wang and Feras Dayoub and Ian Reid},
      year={2025},
      eprint={2501.01163},
      archivePrefix={arXiv},
      primaryClass={cs.CV},
      url={https://arxiv.org/abs/2501.01163}, 
}

@misc{chatterjeeMemoryefficientStreamingVideoLLMs2025,
      title={Memory-efficient Streaming VideoLLMs for Real-time Procedural Video Understanding}, 
      author={Dibyadip Chatterjee and Edoardo Remelli and Yale Song and Bugra Tekin and Abhay Mittal and Bharat Bhatnagar and Necati Cihan Camgöz and Shreyas Hampali and Eric Sauser and Shugao Ma and Angela Yao and Fadime Sener},
      year={2025},
      eprint={2504.13915},
      archivePrefix={arXiv},
      primaryClass={cs.CV},
      url={https://arxiv.org/abs/2504.13915}, 
}

@misc{chenLiveCCLearningVideo2025,
      title={LiveCC: Learning Video LLM with Streaming Speech Transcription at Scale}, 
      author={Joya Chen and Ziyun Zeng and Yiqi Lin and Wei Li and Zejun Ma and Mike Zheng Shou},
      year={2025},
      eprint={2504.16030},
      archivePrefix={arXiv},
      primaryClass={cs.CV},
      url={https://arxiv.org/abs/2504.16030}, 
}

@misc{dingStreamMindUnlockingFull2025,
      title={StreamMind: Unlocking Full Frame Rate Streaming Video Dialogue through Event-Gated Cognition}, 
      author={Xin Ding and Hao Wu and Yifan Yang and Shiqi Jiang and Donglin Bai and Zhibo Chen and Ting Cao},
      year={2025},
      eprint={2503.06220},
      archivePrefix={arXiv},
      primaryClass={cs.CV},
      url={https://arxiv.org/abs/2503.06220}, 
}
}


% \clearpage

% {
%     \small
%     \raggedbottom 
%     \interlinepenalty=0 
%     \bibliographystyle{ieeenat_fullname}
%     \bibliography{main}
% }
\end{document}